\documentclass[11pt,a4paper]{article}
\usepackage{times,latexsym}
\usepackage{url}
\usepackage{comment}
\usepackage[T1]{fontenc}

\usepackage[hyperref]{emnlp2020}
\usepackage{microtype}
\aclfinalcopy 
\def\aclpaperid{***} 

\usepackage{tikz}
\usetikzlibrary{arrows,matrix,graphs,shapes,decorations,automata,backgrounds,petri,positioning,calc,decorations.pathmorphing,external,shapes.multipart}
\tikzstyle{reverseclip}=[insert path={(current page.north east) --
  (current page.south east) --
  (current page.south west) --
  (current page.north west) --
  (current page.north east)}
]

\tikzset{pas/.style={fill=gray!60}, 
act/.style={fill=gray!30},
main/.style={draw,fill=white},
ctx/.style={rounded rectangle,minimum size=7mm},
val/.style={rectangle,minimum size=7mm},
cmd/.style={chamfered rectangle,draw,fill=white},
tns/.style={circle,minimum size=4mm,draw,fill=white},
par/.style={circle,minimum size=4mm,draw,fill=black}, 
minipar/.style={circle,minimum size=2.5mm,draw,fill=black}, 
pn/.style={rounded corners, rectangle,fill=blue!30,draw,minimum size=15mm},
medpn/.style={rounded corners, rectangle,fill=blue!30,draw,minimum size=20mm},
bigpn/.style={rounded corners, rectangle,fill=blue!30,draw,minimum size=25mm}}

\newcommand{\proofspace}{\vphantom{()}}

\usepackage{proof}
\usepackage{amssymb}
\usepackage{pifont}
\usepackage{amsmath}
\usepackage{tabularx}
\usepackage{multirow}
\usepackage{multicol}
\usepackage{diagbox}
\usepackage{bm}
\usepackage{xcolor}
\usepackage{booktabs}
\usepackage{algpseudocode}
\usepackage{algorithm}
\usepackage{float}
\usepackage{rotating}
\usepackage{tikz-dependency}

\newcommand{\li}{\multimap}
\newcommand{\type}[1]{\textsc{#1}}
\newcommand{\term}[1]{\mathtt{#1}}
\renewcommand{\vec}[1]{\bm{#1}}

\newcommand{\n}{\type{n}}
\newcommand{\np}{\type{np}}
\newcommand{\Inf}{\type{inf}}
\newcommand{\CP}{\type{cp}}
\newcommand{\pp}{\type{pp}}
\newcommand{\ww}{\type{ww}}

\newcommand{\ppart}{\type{ppart}}
\newcommand{\pron}{\type{pron}}
\newcommand{\s}{\type{s}_\text{main}}
\newcommand{\ssub}{\type{s}_\text{sub}}

\newcommand{\adj}{\type{adj}}
\newcommand{\dia}[1]{\Diamond^{\text{#1}}}
\newcommand{\bo}[2]{\Box^{\text{#1}}\left({#2}\right)}
\newcommand{\lis}{\!\li\!}
\newcommand{\sos}{\texttt{[SOS]}}
\newcommand{\sep}{\texttt{[SEP]}}
\newcommand{\myraisebox}[1]{\raisebox{-1.8pt}[3pt]{\ensuremath{#1}}}

\newcommand{\duck}{\includegraphics[scale=0.022]{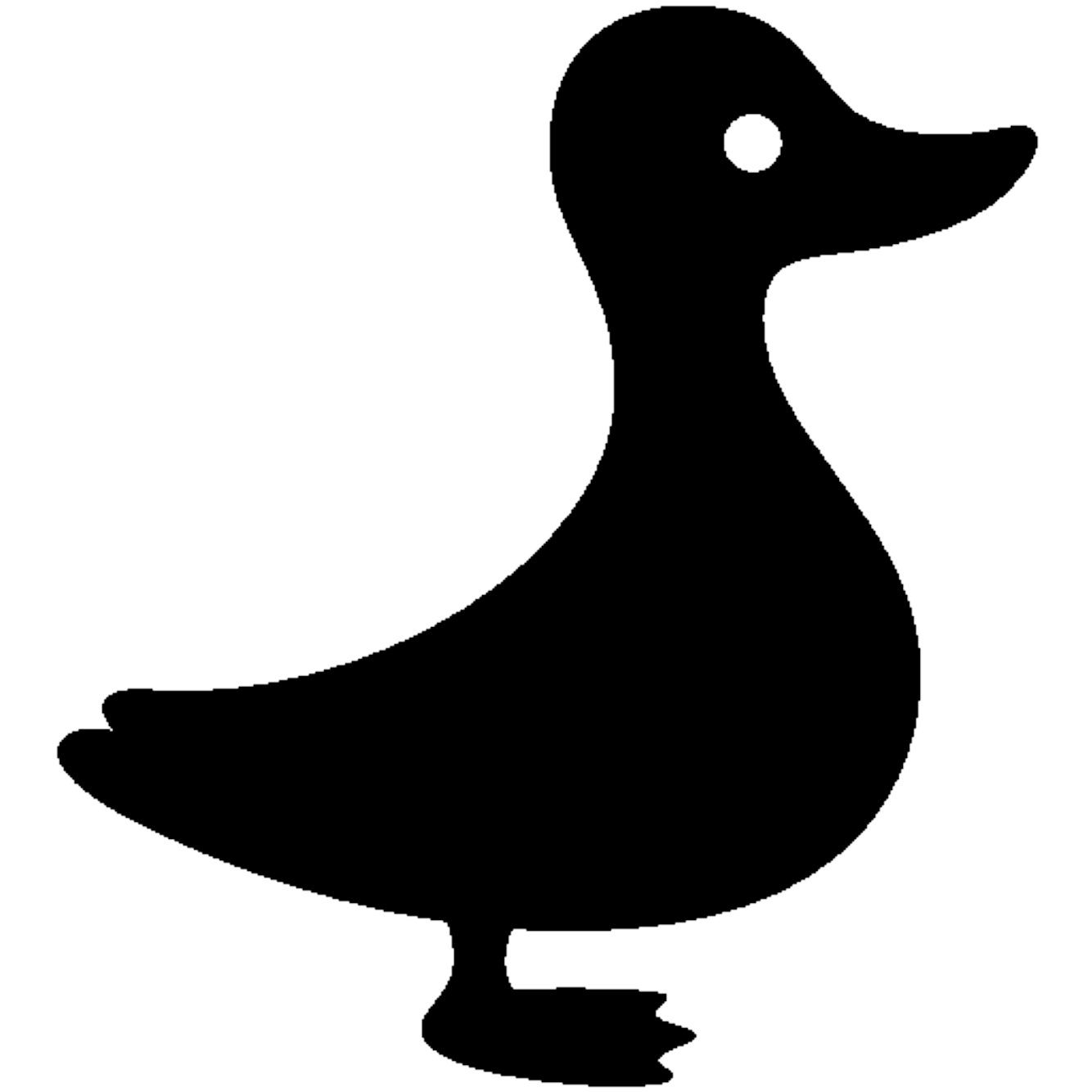}}
\newcommand{\flamingo}{\includegraphics[scale=0.020]{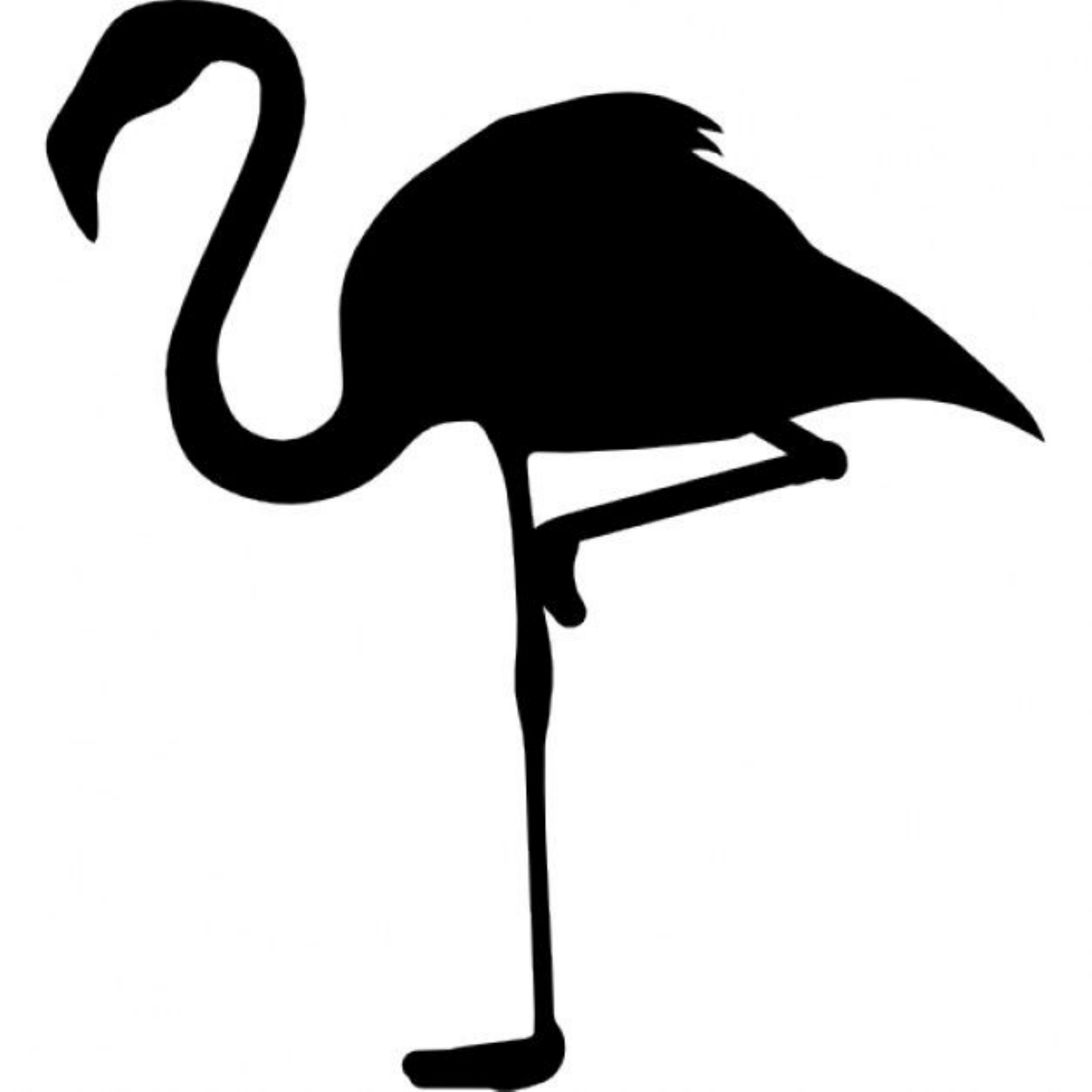}}

\title{Neural Proof Nets}

\author{
    Konstantinos Kogkalidis\textsuperscript{{\duck}} \and
    Michael Moortgat\textsuperscript{\duck} \and
    Richard Moot\textsuperscript{\flamingo} \\ 
    \textsuperscript{\duck} Utrecht Institute of Linguistics OTS, Utrecht University \\ 
    \textsuperscript{\flamingo} LIRMM, Universit\'{e} de Montpellier, CNRS \\
    \texttt{\{k.kogkalidis,m.j.moortgat\}@uu.nl, richard.moot@lirmm.fr}
}

\date{}

\begin{document}
\maketitle
\begin{abstract}
Linear logic and the linear $\lambda$-calculus have a long standing tradition in the study of natural language form and meaning.
Among the proof calculi of linear logic, proof nets are of particular interest, offering an attractive geometric representation of derivations that is unburdened by the bureaucratic complications of conventional prooftheoretic formats.
Building on recent advances in set-theoretic learning, we propose a neural variant of proof nets based on Sinkhorn networks, which allows us to translate parsing as the problem of extracting syntactic primitives and permuting them into alignment.
Our methodology induces a batch-efficient, end-to-end differentiable architecture that actualizes a formally grounded yet highly efficient neuro-symbolic parser.
We test our approach on {\AE}thel, a dataset of type-logical derivations for written Dutch, where it manages to correctly transcribe raw text sentences into proofs and terms of the linear $\lambda$-calculus with an accuracy of as high as $70\%$.
\end{abstract}

\section{Introduction}
There is a broad consensus among grammar formalisms that the composition of form and meaning in natural language
is a resource-sensitive process, with the words making up a phrase contributing exactly once to the resulting whole.
The sentence ``the Mad Hatter offered'' is ill-formed because of a \emph{lack} of grammatical material,
``offer'' being a ditransitive verb; ``the Cheshire Cat grinned Alice a cup of tea'' on the other hand is ill-formed because
of an \emph{excess} of material, which the intransitive verb ``grin'' cannot accommodate.

Given the resource-sensitive nature of language, it comes as no surprise that Linear Logic \cite{girard1987linear}, 
in particular its intuitionistic version ILL, plays a central role in current logic-based grammar formalisms.
Abstract Categorial Grammars and Lambda Grammars~\cite{de2001towards, muskens2001lambda} use ILL ``as-is'' to
characterize an abstract level of grammatical structure from which surface form and semantic interpretation
are obtained by means of compositional translations. Modern typelogical grammars in the tradition of
the Lambek Calculus~\cite{lambek1958mathematics}, e.g.~Multimodal TLG~\cite{Moortgat96}, Displacement Calculus~\cite{Morrill14},
Hybrid TLG~\cite{KubotaLevine20}, refine the type language to account for syntactic aspects 
of word order and constituency; ILL here is the target logic for semantic interpretation, reached by
a homomorphism relating types and derivations of the syntactic calculus to their semantic
counterparts.

A common feature of the aforementioned formalisms is their adoption of the \textit{parsing-as-deduction} method:
determining whether a phrase is syntactically well-formed is seen as the outcome of a process of logical deduction.
This logical deduction automatically gives rise to a program for meaning composition, thanks to the remarkable
correspondence between logical proof and computation known as the Curry-Howard isomorphism~\cite{sorensen2006lectures}, a natural manifestation of the syntax-semantics interface.
The Curry-Howard $\lambda$-terms associated with derivations are neutral with respect to the particular
semantic theory one wants to adopt, accommodating both the truth-conditional view of formal semantics
and the vector-based distributional view~\cite{MuskensSadrzadeh18}, among others.

Despite their formal appeal, grammars based on variants of linear logic have fallen out of
favour within the NLP community, owing to a scarcity of large-scale datasets, but also due to difficulties in aligning them with the established high-performance neural toolkit.
Seeking to bridge the gap between formal theory and applied practice, we focus on the \emph{proof nets} of linear logic, a lean graphical calculus
that does away with the bureaucratic symbol-manipulation overhead characteristic of conventional prooftheoretic presentations (\S\ref{sec:background}).
Integrating proof nets with recent advances in neural processing, we propose a novel approach to linear logic proof search that eliminates issues commonly associated with higher-order types and hypothetical reasoning, while greatly reducing the computational costs of structure manipulation, backtracking and iterative processing that burden standard parsing techniques (\S\ref{sec:npn}).

Our proposed methodology relies on two key components.
The first is an encoder/decoder-based supertagger that converts raw text sentences into linear logic judgements by dynamically constructing contextual type assignments, one primitive symbol at a time.
The second is a bi-modal encoder that contextualizes the generated judgement in conjunction with the input sentence.
The contextualized representations are fed into a Sinkhorn layer, tasked with finding the valid permutation that brings primitive symbol occurrences into alignment.
The architecture induced is trained on labeled data, and assumes the role of a formally grounded yet highly accurate parser, which transforms raw text sentences into linear logic proofs and computational terms of the simply typed linear $\lambda$-calculus, further decorated with dependency annotations that allow reconstruction of the underlying dependency graph (\S\ref{sec:experiments}).

\section{Background}
\label{sec:background}
\begin{figure*}[ht]
\scriptsize
\[
\infer[]{\text{De strategie die ze volgen is eeuwenoud} \vdash 
    \left( \term{is} \ \left(\term{eeuwenoud}\right)^\text{predc} \right)
    \left( \left( \term{die} \ \left(\lambda x^{\text{obj}}.\left( \term{volgen} \ x\right) \left(\term{ze}\right)^{\text{su}} \right)^{\text{body}} \right)_{\text{mod}} \left(\left(\term{de}\right)_\text{det} \ \term{strategie} \right) \right)^\text{su} : 
\s}{
    \hspace{-100pt}
    \infer[]{\myraisebox{\dia{su}\np \lis \s}}{
        \infer[]{\myraisebox{\dia{predc}\adj \lis \dia{su}\np \lis \s}}{\text{is}}
        &
        \infer[]{\myraisebox{\adj}}{\text{eeuwenoud}}
    }
    &
    \infer[]{\myraisebox{\np}}{
        \infer[]{\myraisebox{\bo{mod}{\np\lis\np}}}{
            \infer[]{\myraisebox{\dia{body}\left(\!\dia{obj}\pron\lis\ssub\right)\lis\bo{mod}{\np\lis\np}}}{\text{die}}
            &
            \hspace{-40pt}
            \infer[]{\myraisebox{\dia{obj}\pron\lis\ssub}}{
                \infer[]{\ssub}{
                    \infer[]{\myraisebox{\dia{su}\pron\lis\ssub}}{
                        \infer[]{\myraisebox{\dia{obj}\pron\lis\dia{su}\pron\lis\ssub}}{\text{volgen}}
                        &
                        \hspace{-5pt}
                        \infer[]{\myraisebox{\pron}}{x}
                    }
                    &
                    \hspace{-5pt}
                    \infer[]{\myraisebox{\pron}}{\text{ze}}
                }
            }
        }
        &
        \hspace{-10pt}
        \infer[]{\myraisebox{\np}}{
            \infer[]{\myraisebox{\bo{det}{\n\lis\np}}}{\text{de}}
            &
            \hspace{-5pt}
            \infer[]{\myraisebox{\n}}{\text{strategie}}
        }
    }
}
\]
\caption{Example derivation and Curry-Howard $\lambda$-term for the phrase \textit{De strategie die ze volgen is eeuwenoud} (``The strategy that they follow is ancient'') from \AE thel sample \texttt{dpc-ind-001645-nl-sen.p.12.s.1\_1}, showcasing how hypothetical reasoning enables the derivation of an object-relative clause (note how the instantiation of variable $x$ of type $\pron$ followed by its subsequent abstraction creates an argument for the higher-order function assigned to ``die''). Judgement premises and rule names have been omitted for brevity's sake.}
\label{fig:fish}
\end{figure*}

We briefly summarize the logical background we are assuming, starting with 
ILL${}_{\li}$, the implication-only fragment of ILL, then moving on to the 
dependency-enhanced version ILL${}_{\li,\Diamond,\Box}$ which we employ in our experimental setup.

\subsection{ILL${}_{\li}$}
Formulas (or \textit{types}) of ILL${}_{\li}$ are inductively defined according to the grammar below:
\[
    \mathcal{T} ::= \type{A} \ | \ \type{T}_1 \li \type{T}_2
\]
Formula $\type{A}$ is taken from a finite set of atomic formulas $\mathcal{A} \subset \mathcal{T}$; a complex formula $\type{T}_1 \!\li\! \type{T}_2$ 
is the type signature of a transformation that applies on $\type{T}_1 \in \mathcal{T}$ and produces $\type{T}_2 \in \mathcal{T}$,
consuming the argument in the process. This view of formulas as non-renewable resources makes ILL${}_{\li}$ the
logic of \emph{linear} functions.\footnote{We refer to ~\citet{wadler1993taste} for a gentle introduction.}

We can present the inference rules of ILL${}_{\li}$ together with the associated linear $\lambda$-terms in Natural
Deduction format. Judgements are sequents of the form $\term{x}_1:\type{T}_1,\ldots,\term{x}_n:\type{T}_n \vdash \term{M}:\type{C}$.
The antecedent left of the turnstile is a \emph{typing environment} (or \emph{context}), a sequence of variables $\term{x}_i$, each given a type declaration $\type{T}_i$.
These variables serve as the \emph{parameters} of a program $\term{M}$ of type $\type{C}$ that corresponds to the proof
of the sequent.

Proofs are built from axioms $\term{x}:\type{T} \vdash \term{x}:\type{T}$ with the aid of two rules of inference:
\begin{equation}
\label{eqn:elimination}
\infer[\li  E]{\Gamma,\Delta\vdash \term{(M\ N)}: \type{T}_2}{\Gamma\vdash \term{M}: \type{T}_1\li \type{T}_2 & \Delta\vdash \term{N}: \type{T}_1}
\end{equation}

\begin{equation}
\label{eqn:introduction}
    \infer[\li I]{\Gamma \vdash \lambda\term{x.M}: \type{T}_1\li \type{T}_2}{\Gamma, \term{x}: \type{T}_1 \vdash \term{M}: \type{T}_2}\\
\end{equation}

(\ref{eqn:elimination}) is the elimination of the implication and models \textit{function application};
it proposes that if from some context $\Gamma$ one can derive a program $\term{M}$ of type $\type{T}_1 \!\li\! \type{T}_2$,
and from context $\Delta$ one can derive a program $\term{N}$ of type $\type{T}_1$, then from the multiset union $\Gamma, \Delta$ one
can derive a term $\term{(M\ N)}$ of type $\type{T}_2$.

(\ref{eqn:introduction}) is the introduction of the implication and models \textit{function abstraction};
it proposes that if from a context $\Gamma$ together with a type declaration $\term{x}:\type{T}_1$ one can derive a program
term $\term{M}$ of type $\type{T}_2$, then from $\Gamma$ alone one can derive the abstraction $\term{\lambda x.M}$,
denoting a linear function of type $\type{T}_1\!\li\!\type{T}_2$.

To obtain a \emph{grammar} based on ILL$_{\li}$, we consider the logic in combination with
a \textit{lexicon}, assigning one or more type formulas to the words of the language. In this setting, the proof
of a sequent $\term{x}_1:\type{T}_1,\ldots,\term{x}_n:\type{T}_n \vdash \term{M}:\type{C}$ constitutes an algorithm
to compute a meaning $\term{M}$ of type $\type{C}$,
given by substituting parameters $\term{x}_i$ with lexical meanings $\term{w}_i$.
In the type lexicon, atomic types are used to denote syntactically autonomous, stand-alone units (words and phrases);
e.g.~$\np$ for noun-phrase, $\type{s}$ for sentence, etc. Function types are assigned to incomplete expressions,
e.g. $\np\!\li\!\type{s}$
for an intransitive verb consuming a noun-phrase to produce a sentence, $\np\!\li\np\!\li\type{s}$ for a transitive verb, etc.\footnote{Read $\li$ as right-associative.}
Higher-order types, i.e.~types of order greater than 1, denote functions that apply to functions; these give the grammar access to hypothetical reasoning, in virtue of the implication introduction rule.\footnote{$\mathcal{O}(\type{A})$, the order of an atomic type, equals zero;
for function types $\mathcal{O}(\type{T}_1\!\li\!\type{T}_2) = \textrm{max}(\mathcal{O}(\type{T}_1) + 1, \mathcal{O}(\type{T}_2))$.}
Combined with parametric polymorphism, higher-order types eschew the need for phantom syntactic nodes, enabling straightforward derivations for apparent non-linear phenomena involving long-range dependencies, elliptical conjunctions, wh-movement and the like.

\newcommand{\posimpl}{\overset{+}{\multimap}}
\newcommand{\negimpl}{\overset{-}{\multimap}}
\newcommand{\posat}[1]{\overset{+}{#1\proofspace}}
\newcommand{\negat}[1]{\overset{-}{#1\proofspace}}
\begin{figure*}[ht]
\scriptsize
\begin{center}
\begin{tikzpicture}
\node (the) at (-8em,0em) {$\overset{+}{A\multimap B}$};
\node (nposa) at (-10em,4em) {$\negat{A}$};
\draw (the) -- (nposa);
\node (thenp) at (-6em,4em) {$\posat{B}$};
\draw (the) -- (thenp);
\node (aplus) at (2em,0em) {$\overset{+}{A}$};
\node (amin) at (8em,0em) {$\overset{-}{A}$};
\draw (aplus) -- (2em,2em) -- (8em,2em) -- (amin); 
\node (the) at (20em,0em) {$\overset{-}{A\multimap B}$};
\node (nposa) at (18em,4em) {$\posat{A}$};
\draw [dashed] (the) -- (nposa);
\node (thenp) at (22em,4em) {$\negat{B}$};
\draw [dashed] (the) -- (thenp);
\end{tikzpicture}
\end{center}
\caption{Links for linear logic proof nets. Left/right: positive/negative implication. Center: axiom link.}
\label{fig:links}
\end{figure*}
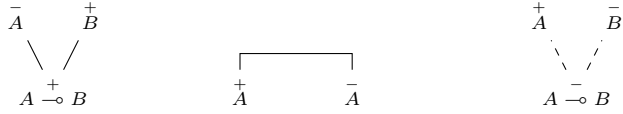

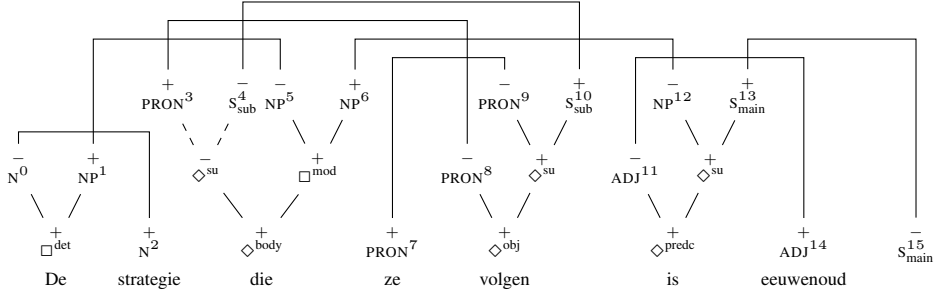
\begin{figure*}[ht]
\scriptsize
\begin{center}
\begin{tikzpicture}
\node (de) at (0em, 0em) {\proofspace\text{De}};
\node (dedet) at (0em, 2em) {$\posat{\Box^{\text{det}}}$};
\node (den) at (-2em, 6em) {$\negat{\n^0}$};
\node (denp) at (2em, 6em) {$\posat{\np^1}$};
\draw (dedet) -- (den);
\draw (dedet) -- (denp);
\node (strategie) at (5em, 0em) {\proofspace\text{strategie}};
\node (strategien) at (5em, 2em) {$\posat{\n^2}$};
\node (die) at (11em, 0em) {\proofspace\text{die}};
\node (diebody) at (11em, 2em) {$\posat{\Diamond^{\text{body}}}$};
\node (dieobj) at (8em, 6em) {$\negat{\Diamond^{\text{su}}}$};
\node (diemod) at (14em, 6em) {$\posat{\Box^{\text{mod}}}$};
\node (diepron) at (6em, 10em) {$\posat{\pron^3}$};
\node (diessub) at (10em, 10em) {$\negat{\ssub^4}$};
\node (dienpneg) at (12em, 10em) {$\negat{\np^5}$};
\node (dienppos) at (16em, 10em) {$\posat{\np^6}$};
\draw (diebody) -- (dieobj);
\draw (diebody) -- (diemod);
\draw [dashed] (dieobj) -- (diepron);
\draw [dashed] (dieobj) -- (diessub);
\draw (diemod) -- (dienpneg);
\draw (diemod) -- (dienppos);
\node (ze) at (18em, 0em) {\proofspace\text{ze}};
\node (zepron) at (18em, 2em) {$\posat{\pron^7}$};
\node (volgen) at (24em, 0em) {\proofspace\text{volgen}};
\node (volgenobj) at (24em, 2em) {$\posat{\Diamond^{\text{obj}}}$};
\node (volgenpronobj) at (22em, 6em) {$\negat{\pron^8}$};
\node (volgensu) at (26em, 6em) {$\posat{\Diamond^{\text{su}}}$};
\node (volgenpronsu) at (24em, 10em) {$\negat{\pron^9}$};
\node (volgenssub) at (28em, 10em) {$\posat{\ssub^{10}}$};
\draw (volgenobj) -- (volgenpronobj);
\draw (volgenobj) -- (volgensu);
\draw (volgensu) -- (volgenpronsu);
\draw (volgensu) -- (volgenssub);
\node (is) at (33em, 0em) {\proofspace\text{is}};
\node (ispredc) at (33em, 2em) {$\posat{\Diamond^{\text{predc}}}$};
\node (isadj) at (31em, 6em) {$\negat{\adj^{11}}$};
\node (issu) at (35em, 6em) {$\posat{\Diamond^{\text{su}}}$};
\node (isnp) at (33em, 10em) {$\negat{\np^{12}}$};
\node (iss) at (37em, 10em) {$\posat{\s^{13}}$};
\draw (ispredc) -- (isadj);
\draw (ispredc) -- (issu);
\draw (issu) -- (isnp);
\draw (issu) -- (iss);
\node (eeuwenoud) at (40em, 0em) {\proofspace\text{eeuwenoud}};
\node (eeuwenoudadj) at (40em, 2em) {$\posat{\adj^{14}}$};
\node (goal) at (46em, 2em) {$\negat{\s^{15}}$};
\draw (den) -- (-2em, 8em) -- (5em, 8em) -- (strategien);
\draw (denp) -- (2em, 13em) -- (12em, 13em) -- (dienpneg);
\draw (dienppos) -- (16em, 13em) -- (33em, 13em) -- (isnp);
\draw (isadj) -- (31em, 12em) -- (40em, 12em) -- (eeuwenoudadj);
\draw (iss) -- (37em, 13em) -- (46em, 13em) -- (goal);
\draw (zepron) -- (18em, 12em) -- (24em, 12em) -- (volgenpronsu);
\draw (volgenpronobj) -- (22em, 14em) -- (6em, 14em) -- (diepron);
\draw (diessub) -- (10em, 15em) -- (28em, 15em) -- (volgenssub);
\end{tikzpicture}
\end{center}
\caption{Proof net corresponding to the natural deduction derivation of Figure~\ref{fig:fish}, with modal markings in place of implication arrows.
Atomic types at the fringe of the formula decomposition trees are marked with superscript indices denoting their position for ease of identification.
During decoding, the proof frame is flattened as the linear sequence:
\small{
$\left[ \sos, \Box^{\text{det}}, \n, \np, \sep, \n, \sep, \dia{body}, \dia{su}, \pron,\\ \ssub, \Box^\text{mod}, \np, \np, \sep, \pron, \sep, \dia{obj}, \dots \right]$
}
}
\label{fig:pn}
\end{figure*}

\subsection{ILL$_{\li,\Diamond,\Box}$}
\label{subsec:deps}
For our experimental setup, we will be utilizing the \AE thel dataset, a Dutch corpus of type-logical derivations~\cite{kogkalidis-moortgat-moot:2020:LREC}.
Non-commutative categorial grammars in the tradition of~\citet{lambek1958mathematics} attempt to directly capture syntactic fine-structure by making a distinction between left- and right-directed variants of the implication.
In order to deal with the relatively free word order of Dutch and contrary to the former, \AE thel's type system sticks to the directionally non-committed $\li$ for function types, but
compensates with two strategies for introducing syntactic discrimination. First, the \emph{atomic} type inventory distinguishes
between major clausal types $\type{s}_\text{sub}, \type{s}_\text{v1}, \type{s}_\text{main}$, based
on the positioning of their verbal head (clause final, clause initial, verb second, respectively).
Secondly, \emph{function} types are enhanced with dependency information, expressed via
a family of unary modalities $\Diamond^{d}$, $\Box^{m}$, with dependency labels $d,m$ drawn from 
disjoint sets of complement vs adjunct markers.
The new constructors produce types $\Diamond^d \type{A}\!\li\!\type{B}$, used to denote the \emph{head} of a phrase $\type{B}$ that selects for a \emph{complement} $\type{A}$ and assigns it the dependency role $d$, and types $\Box^m (\type{A}\!\li\!\type{B})$, used to denote \emph{adjuncts}, i.e.~non-head functions that project the dependency role $m$ upon application. Following dependency grammar tradition, determiners and modifiers are treated as non-head functions.

The type enhancement induces a dependency marking on the derived $\lambda$-term, reflecting the introduction/elimination of the $\Diamond,\Box$ constructors; each dependency domain has a unique head, together with its complements and possible adjuncts, denoted by superscripts and subscripts, respectively.
Figure~\ref{fig:fish} provides an example derivation and the corresponding $\lambda$-term.

A shallow dependency graph can be trivially reconstructed by traversal of the decorated $\lambda$-term, recursively establishing labeled edges along the path from a phrasal head to the head of each of its dependants while skipping abstractions; see Figure~\ref{fig:depgraph} for an example.

\begin{figure}[H]
    \centering
    \begin{dependency}[theme=simple]
        \begin{deptext}
            De \& strategie \& die \& ze \& volgen \& is \& eeuwenoud \\
        \end{deptext}
        \deproot{6}{ROOT}
        \depedge{6}{7}{predc}
        \depedge{6}{2}{su}
        \depedge{2}{1}{det}
        \depedge{2}{3}{mod}
        \depedge{3}{5}{body}
        \depedge{5}{4}{su}
    \end{dependency}
    \caption{Shallow graph for the term of Figure~\ref{fig:fish}.}
    \label{fig:depgraph}
\end{figure}

\subsection{Proof Nets}
\label{subsec:pn}
Despite their clear computational interpretation~\cite{glt, bpt, sorensen2006lectures}, proofs in natural deduction format are arduous to obtain; reasoning with hypotheticals necessitates a mixture of forward and backward chaining search strategies.
The sequent calculus presentation, on the other hand, permits exhaustive proof search via pure backward chaining, but does so at the cost of spurious ambiguity.
Moreover, both the above assume a tree-like proof structure, which hinders their parallel processing and impairs compatibility with neural methods.
As an alternative, we turn our attention towards \textit{proof nets}~\cite{girard1987linear}, a graphical representation of linear logic proofs that captures hypothetical reasoning in a purely geometric manner.
Proof nets may be seen as a parallelized version of the sequent calculus or a multi-conclusion version of natural deduction and combine the best of both words, allowing for 
flexible and easily parallelized 
 proof search while maintaining the 1-to-1 correspondence with the terms of the linear $\lambda$-calculus.

To define ILL proof nets, we first need the auxiliary notion of \emph{polarity}. 
We assign \emph{positive} polarity to resources we have, \emph{negative} polarity to resources we seek.
Logically, a formula with negative polarity appears in \textit{conclusion} position (right of the turnstile), whereas formulas with positive polarity appear in \textit{premise} position (left of the turnstile).
Given a formula and its polarity, the polarity of its subformulas is computed as follows: for a positive formula $\type{T}_1\!\li\!\type{T}_2$, $\type{T}_1$ is negative and $\type{T}_2$ is positive, whereas for a negative formula $\type{T}_1\!\li\!\type{T}_2$, $\type{T}_1$ is positive and $\type{T}_2$ is negative. 

With respect to proof search, proof nets present a simple but general setup as follows.
(1) Begin by writing down the formula decomposition tree for all formulas in a sequent $\type{P}_1,\ldots \type{P}_n \vdash \type{C}$, keeping track of polarity information; the result is called a \emph{proof frame}.
(2) Find a perfect matching between the positive and negative atomic formulas; the result is called a \emph{proof structure}.
(3) Finally, verify that the proof structure satisfies the correctness condition; if so, the result is a \emph{proof net}.

Formula decomposition is fully deterministic, with the decomposition rules shown in Figure~\ref{fig:links}. 
There are two logical links, denoting positive and negative occurrences of an implication (corresponding to the elimination and introduction rules of natural deduction, respectively). 
A third rule, called the axiom link, connects two equal formulas of opposite polarity.

To transform a proof frame into a proof structure, we first need to check the \textit{count invariance} property, which requires an equal count of positive and negative occurrences for every atomic type, and then connect atoms of opposite polarity.
In principle, we can connect any positive atom to any negative atom when both are of the same type; the combinatorics of proof search lies, therefore, in the axiom connections (the number of possible proof structures scales factorial to the number of atoms).
Not all proof structures are, however, proof nets.
Validating the correctness of a proof net can be done in linear time~\cite{pnlinear, murong}; a common approach is to attempt a traversal of the proof net, ensuring that all nodes are visited (connectedness) and no loops exist (acyclicity)~\cite{multiplicatives}.
There is an apparent tension here between finding just \emph{a} matching of atomic formulas (which is trivial once we satisfy the count invariance)
and finding \emph{the} correct matching, which produces not only a proof net, but also the preferred semantic reading of the sentence. 
Deciding the provability of a linear logic sequent is an NP-complete problem~\cite{lincoln}, even in the simplest case where formulas are restricted to order 1~\cite{Kanovich94}.
Figure~\ref{fig:pn} shows the proof net equivalent to the derivation of Figure~\ref{fig:fish}.

\section{Neural Proof Nets}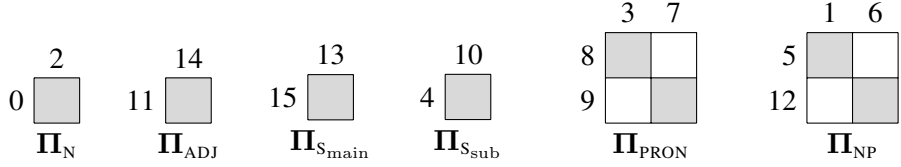
\begin{figure*}[ht]
    \[
    \begin{tikzpicture}[scale=.6]
    \fill[gray!30] (0,0) rectangle +(1,1);
    \node [above] at (0.5,1) {2};
    \node [left] at (0,0.5) {0};
    \node [below] at (.5,0) {$\vec{\Pi}_\n$};
    \draw (0,0) grid (1,1);
    \end{tikzpicture}
    \quad
    \begin{tikzpicture}[scale=.6]
    \fill[gray!30] (0,0) rectangle +(1,1);
    \node [above] at (0.5,1) {14};
    \node [left] at (0,0.5) {11};
    \node [below] at (.5,0) {$\vec{\Pi}_\type{adj}$};
    \draw (0,0) grid (1,1);
    \end{tikzpicture}
    \quad
    \begin{tikzpicture}[scale=.6]
    \fill[gray!30] (0,0) rectangle +(1,1);
    \node [above] at (0.5,1) {13};
    \node [left] at (0,0.5) {15};
    \node [below] at (.5,0) {$\vec{\Pi}_{\type{s}_\mathrm{main}}$};
    \draw (0,0) grid (1,1);
    \end{tikzpicture}
    \quad
    \begin{tikzpicture}[scale=.6]
    \fill[gray!30] (0,0) rectangle +(1,1);
    \node [above] at (0.5,1) {10};
    \node [left] at (0,0.5) {4};
    \node [below] at (.5,0) {$\vec{\Pi}_{\type{s}_\mathrm{sub}}$};
    \draw (0,0) grid (1,1);
    \end{tikzpicture}
    \qquad
    \begin{tikzpicture}[scale=.6]
    \fill[gray!30] (0,1) rectangle +(1,1);
    \fill[gray!30] (1,0) rectangle +(1,1);
    \node [above] at (0.5,2) {3};
    \node [above] at (1.5,2) {7};
    \node [left] at (0,1.5) {8};
    \node [left] at (0,0.5) {9};
    \node [below] at (1,0) {$\vec{\Pi}_\pron$};
    \draw (0,0) grid (2,2);
    \end{tikzpicture}
    \qquad
    \begin{tikzpicture}[scale=.6]
    \fill[gray!30] (0,1) rectangle +(1,1);
    \fill[gray!30] (1,0) rectangle +(1,1);
    \node [above] at (0.5,2) {1};
    \node [above] at (1.5,2) {6};
    \node [left] at (0,1.5) {5};
    \node [left] at (0,0.5) {12};
    \node [below] at (1,0) {$\vec{\Pi}_\np$};
    \draw (0,0) grid (2,2);
    \end{tikzpicture}
    \]
    \caption{An alternative view of the axiom links of Figure~\ref{fig:pn}, with tables $\vec{\Pi}_\type{n}$, $\vec{\Pi}_\adj$, $\vec{\Pi}_{\s}$, $\vec{\Pi}_{\ssub}$, $\vec{\Pi}_{\type{pron}}$, $\vec{\Pi}_\np$ depicting the linked indices and corresponding permutations for each atomic type in the sentence.}
    \label{fig:tables}
\end{figure*}

\label{sec:npn}
To sidestep the complexity inherent in the combinatorics of linear logic proof search, we investigate proof net construction from a neural perspective.
First, we will need to convert a sentence into a proof frame, i.e. the decomposition of a logical judgement of the form $\type{P}_1,\ldots \type{P}_n \vdash \type{C}$, with $\type{P}_i$ the type of word $i$ and $\type{C}$ the goal type to be derived.
Having obtained a correct proof frame, the problem boils down to establishing axiom links between the set of positive and negative atoms and verifying their validity according to the correctness criteria.
We address each of these steps via a functionally independent neural module, and define \textit{Neural Proof Nets} as their composition.

\subsection{Proof Frames}
\label{subsec:pf}
Obtaining proof frames is a special case of supertagging, a common problem in NLP literature~\cite{bangalore1999supertagging}.
Conventional practice treats supertagging as a discriminative sequence labeling problem, with a neural model contextualizing the tokens of an input sentence before passing them through a linear projection in order to convert them to class weights~\cite{xu2015ccg, vaswani2016supertagging}.
Here, instead, we adopt the generative paradigm~\cite{kogkalidis2019constructive,bhargava-penn-2020-supertagging}, whereby each type is itself perceived as a sequence of primitive symbols.

Concretely, we perform a depth-first-left-first traversal of formula trees to convert types to prefix (Polish) notation.
This converts a type to a linear sequence of symbols $s\! \in \! \mathcal{V}$, where $\mathcal{V}$\! =\! $\mathcal{A}\cup\mathcal{D}$, the union of atomic types and dependency-decorated modal markings.\footnote{Dependency decorations occur only within the scope of an implication, so the two are merged into a single symbol for reasons of length economy.}
Proof frames can then be represented by joining individual type representations, separated with an extra-logical token $\texttt{[SEP]}$ denoting type breaks and prefixed with a special token $\texttt{[SOS]}$ to denote the sequence start (see the caption of Figure~\ref{fig:pn} for an example).
The resulting sequence becomes the goal of a decoding process conditional on the input sentence, as implemented by a sequence-to-sequence model.

Treating supertagging as auto-regressive decoding enables the prediction of any valid type in the grammar, improving generalization and eliminating the need for a strictly defined type lexicon.
Further, the decoder's  comprehension  of  the  type  construction  process  can  yield  drastic  improvements  for beam  search,  allowing  distinct  branching  paths within individual types.
Most importantly, it grants access to the atomic sub-formulas of a sequent, i.e. the primitive entities to be paired within a proof net -- a quality that will come into play when considering the axiom linking process later on.
 
\subsection{Proof Structures}
The conversion of a proof frame into a proof structure requires establishing a correct bijection between positive and negative atoms, i.e. linking each positive occurrence of an atom with a single unique negative occurrence of the same atom.

We begin by first noting that each atomic formula occurrence within a proof frame can be assigned an identifying index according to its position in the sequence (refer to the example of Figure~\ref{fig:pn}).
For each distinct atomic type, we can then create a table with rows enumerating negative and columns enumerating positive occurrences of that type, ordered by their indexes.
We mark cells indexing linked occurrences and leave the rest empty; tables for our running example can be seen in Figure~\ref{fig:tables}.
The resulting tables correspond to a \textit{permutation matrix} $\vec{\Pi}_\type{A}$ for each atomic type $\type{A}$, i.e. a set of matrices that are square, binary and doubly-stochastic, encoding the permutation over the \textit{chain} (i.e. ordered set) of negative elements that aligns them with the chain of matching positive elements.
This key insight allows us to reframe automated proof search as learning the latent space that dictates the permutations between disjoint and non-contiguous sub-sequences of the primitive symbols constituting a decoded proof frame.

Permutation matrices are discrete mathematical objects that are not directly attainable by neural models.
Their continuous relaxations are, however, valid outputs, approximated by means of the Sinkhorn operator~\cite{sinkhorn1964relationship}.
In essence, the operator and its underlying theorem state that the iterative normalization (alternating between rows and columns) of a square matrix with positive entries yields, in the limit, a doubly-stochastic matrix, the entries of which are \textit{almost} binary.
Put differently, the Sinkhorn operator gives rise to a non-linear activation function that applies on matrices, pushing them towards binarity and bistochasticity, analogous to a 2-dimensional softmax that preserves assignment~\cite{mena2018learning}.
Moving to the logarithmic space eliminates the positive entry constraint and facilitates numeric stability through the log-sum-exp trick.
In that setting, the Sinkhorn-normalization of a real-valued square matrix $\vec{X}$ is defined as:
$$\mathrm{Sinkhorn}(\vec{X})= \lim_{\tau \to \infty}\mathrm{exp}\left(\mathrm{Sinkhorn}^\tau(\vec{X})\right)$$ where the induction is given by:\begin{align*}
    \mathrm{Sinkhorn}^0(\vec{X}) &= \vec{X} \\
    \mathrm{Sinkhorn}^\tau(\vec{X}) &= \mathcal{T}_r\left(\mathcal{T}_r\left(\mathrm{Sinkhorn}^{(\tau-1)}(\vec{X})\right)^\top\right)
\end{align*}%
with $\mathcal{T}_r$ the row normalization in the log-space:
\[
{\mathcal{T}_r(\vec{X})}_{i, j} = \vec{X}_{i, j} - log\sum\limits_{r=0}^{N-1}{e^{(\vec{X}_{r, j} - \mathrm{max}(\vec{X}_{r, :}))}}
\]

Bearing the above in mind, our goal reduces to assembling a matrix for each atomic type in a proof frame, with entries containing the unnormalized agreement scores of pairs in the cartesian product of positive and negative occurrences of that type.
Given contextualized representations for each primitive symbol within a proof frame, scores can be simply computed as the inter-representation \textit{dot-product attention}.
Assuming, for instance, $\vec{I}^+_\type{A}$ and $\vec{I}^-_\type{A}$ the vectors indexing the positions of all $a$ positive and negative occurrences of type $\type{A}$ in a proof frame sequence, we can arrange the matrices $\vec{P}_\type{A}, \vec{N}_\type{A} \in \mathbb{R}^{a\times d}$ containing their respective contextualized $d$-dimensional representations (recall that the count invariance property asserts equal shapes).
The dot-product attention matrix containing their element-wise agreements will then be given as $\vec{\tilde{S}}_\type{A} = \vec{P}_\type{A}\vec{N}_\type{A}^\top \in \mathbb{R}^{a \times a}$.
Applying the Sinkhorn operator, we obtain $\vec{S}_\type{A} = \mathrm{Sinkhorn}(\vec{\tilde{S}}_\type{A})$, which, in our setup, will be modeled as a continuous approximation of the underlying permutation matrix $\vec{\Pi}_\type{A}$.

\subsection{Implementation}
\paragraph{Encoder-Decoder}
We first encode sentences using BERTje~\cite{de2019bertje}, a pretrained BERT-Base model~\cite{devlin2018bert} localized for Dutch.
We then decode into proof frame sequences using a Transformer-like decoder~\cite{vaswani2017attention}.

\paragraph{Symbol Embeddings}
In order to best utilize the small, structure-rich vocabulary of the decoder, we opt for lower-dimensional, position-dependent symbol embeddings.
We follow insights from~\citet{Wang2020Encoding} and embed decoder symbols as continuous functions in the complex space, associating each output symbol $s\in\!\mathcal{V}$ with a magnitude embedding $\vec{r_s} \! \in\! \mathbb{R}^{128}$ and a frequency embedding $\vec{\omega_s} \!\in\! \mathbb{R}^{128}$.
A symbol $s$ occurring in position $p$ in the proof frame is then assigned a vector $\vec{\tilde{v}_{s,p}} \! = \! \vec{r_s} e^{j\vec{\omega_s}p}\! \in\! \mathbb{C}^{128}$.
We project to the decoder's vector space by concatenating the real and imaginary parts, obtaining the final representation as $\vec{v_{s,p}} = \mathrm{conc}(\Re(\vec{\tilde{v}_{s,p}}), \Im(\vec{\tilde{v}_{s,p}})) \in \mathbb{R}^{256}$.

Tying the embedding parameters with those of the pre-softmax transformation reduces the network's memory footprint and improves representation quality~\cite{press2017using}.
In duality to the input embeddings, we treat output embeddings as functionals parametric to positions.
To classify a token occurring in position $p$, we first compute a matrix $\vec{V_p}$ consisting of the local embeddings of all vocabulary symbols, $\vec{V_p} = \vec{v_{:, p}} \in \mathbb{R}^{||\mathcal{V}||\times 256}$.
The transpose of that matrix acts then as a linear map from the decoder's representation to class weights, from which a probability distribution is obtained by application of the softmax function.

\paragraph{Proof Frame Contextualization}
Proof frames may generally give rise to more than one distinct proof, with only a portion of those being linguistically plausible.
Frames eligible to more than one potential semantic reading can be disambiguated by accounting for statistical preferences, as exhibited by lexical cues.
Consequently, we need our contextualization scheme to incorporate the sentential representation in its processing flow.
To that end, we employ another Transformer decoder, now modified to operate with no causal mask, thus allowing all decoded symbols to freely attend over one another regardless of their relative position.
This effectively converts it into a \textit{bi-modal encoder} which operates on two input sequences of different length and dimensionality, namely the BERT output and the sequence of proof frame symbols, and constructs contextualized representations of the latter as informed by the former.

\paragraph{Axiom Linking}
We index the contextualized proof frame to obtain a pair of matrices for each distinct atomic type in a sentence, easing the complexity of the problem by preemptively dismissing the possibility of linking unequal types; this also alleviates performance issues noted when permuting sets of high cardinality~\cite{mena2018learning}.
Post contextualization, positive and negative items are projected to a lower dimensionality via a pair of feed-forward neural functions, applied token-wise.
Normalizing the dot-product attention weights between the above with Sinkhorn yields our final output.

\begin{table*}
    \centering
    \newcolumntype{C}{>{\centering\arraybackslash}X}
    \newcolumntype{L}{>{\raggedright\arraybackslash}p{0.28\textwidth}}
    \newcolumntype{R}{>{\raggedleft\arraybackslash}p{0.28\textwidth}}
    \newcommand{\ra}[1]{\renewcommand{\arraystretch}{#1}}
    \ra{1.1}
    \begin{tabularx}{0.95\textwidth}{@{}L|CCCCC|C@{}}
    \multirow{2}{*}{\textbf{Metric} (\%)} & \multicolumn{5}{c|}{\textbf{Beam Size} $\vec\beta$} & \multicolumn{1}{c}{\textbf{Baseline}}\\ 
    & $\beta=1$ & $\beta=2$ & $\beta=3$ & $\beta=5$ & {$\beta=7$} & {\textit{alpino}}\\
    \hline  
    \hline
    \multicolumn{1}{R|}{\textbf{Token Level}} & \multicolumn{5}{c|}{} & \multicolumn{1}{c}{} \\
    Types Correct & 85.5 & 91.4 & 92.4 & 93.2 & 93.4 & 56.2\\
    \multicolumn{1}{R|}{\textbf{Sentence Level}} & \multicolumn{5}{c|}{} & \multicolumn{1}{c}{} \\
    Invariance Correct & 87.6 & 93.4 & 94.9 & 96.1 & 96.6 & \textit{n/a} \\
    Frame Correct & 57.6 & 65.3 & 68.0 & 69.6 & 70.2 & \textit{n/a}\\
    Term Correct \footnotesize{(w/o types)} & 60.0 & 65.6 & 67.7 & 69.1 & 69.6 & 45.7 \\
    Term Correct \footnotesize{(/w types \& deps)} & 56.9 & 63.7 & 65.9 & 67.1 & 67.6 & 30.4
    \end{tabularx}
    \caption{Test set model performance broken down by beam size, and baseline comparison.}
    \label{tab:numbers}
\end{table*}

\section{Experiments}
\label{sec:experiments}
We train, validate and test our architecture on the corresponding subsets of the \AE thel dataset, filtering out samples the proof frames of which exceed $100$ primitive symbols.
Implementation details and hyper-parameter tables, an illustration of the full architecture, dataset statistics and example parses are provided in Appendix~\ref{sec:appendix}.\footnote{The implementing code can be found at \url{github.com/konstantinosKokos/neural-proof-nets}.}

\subsection{Training}
We train our architecture end-to-end, including all BERT parameters apart from the embedding layer, using AdamW~\cite{loshchilov2018fixing}.

In order to jointly learn representations that accommodate both the proof-frame and the proof-structure outputs, we back-propagate a loss signal derived as the addition of two loss functions.
The first is the Kullback-Leibler divergence between the predicted proof frame symbols and the label-smoothed ground-truth distribution~\cite{muller2019does}.
The second is the negative log-likelihood between the Sinkhorn-activated dot-product weights and the corresponding binary-valued permutation matrices.

Throughout training, we validate by measuring the per-symbol and per-sentence typing accuracy of the greedily decoded proof frame, as well as the linking accuracy under the assumption of an error-free decoding.
We perform model selection on the basis of the above metrics and reach convergence after approximately 300 epochs.

\subsection{Testing}
\label{subsec:test}
We test model performance using beam search.
For each input sentence, we consider the $\beta$ best decode paths, with a path's score being the sum of its symbols' log probabilities, counting all symbols up to the last expected \texttt{[SEP]} token.
Neural decoding is followed by a series of filtering steps.
We first parse the decoded symbol sequences, discarding beams containing subsequences that do not meet the inductive constructors of the type grammar.
The atomic formulas of the passing proof frames are polarized according to the process of ~\S\ref{subsec:pn}.
Frames failing to satisfy the count invariance property are also discarded.
The remaining ones constitute potential candidates for a proof structure;
their primitive symbols are contextualized by the bimodal encoder, and are then used to compute soft axiom link strengths between atomic formulas of matching types.
Discretization of the output yields a graph encoding a proof structure;
we follow the net traversal algorithm of~\citet{lamarcheEssential} to check whether it is a valid proof net, and, if so,  produce the $\lambda$-term in the process~\cite{degrooteRetore}.
Terms generated this way contain no redundant abstractions, being in $\beta$-normal $\eta$-long form.

\subsection{Analysis}
Table~\ref{tab:numbers} presents a breakdown of model performance at different beam widths.
To evaluate model performance, we use the first valid beam of each sample, defaulting to the highest scoring beam if none is available.
On the token level, we report \textit{supertagging accuracy}, i.e. the percentage of types correctly assigned.
We further measure the percentage of samples satisfying each of the following sentential metrics:
1) \textit{invariance property}, a condition necessary for being eligible to a proof structure,
2) \textit{frame correctness}, i.e. whether the decoded frame is identical to the target frame, meaning all types assigned are the correct ones,
3) \textit{untyped term accuracy}, i.e. whether, regardless of the proof frame, the untyped $\lambda$-term coincides with the true one,
and 4) \textit{typed term accuracy}, meaning that both the proof frame and the untyped term are correct.

Numeric comparisons against other works in the literature is neither our prime goal nor an easy task; the dataset utilized is fairly recent, the novelty of our methods renders them non-trivial to adapt to other settings, and ILL-friendly categorial grammars are not particularly common in experimental setups.
As a sanity check, however, and in order to obtain some meaningful baselines, we employ the Alpino parser~\cite{bouma2001alpino}.
Alpino is a hybrid parser based on a sophisticated hand-written grammar and a maximum entropy disambiguation model; despite its age and the domain difference, Alpino is competitive to the state-of-the-art in UD parsing, remaining within a 2\% margin to the last reported benchmark~\cite{bouma-van-noord-2017-increasing,che-etal-2018-towards}.
We pair Alpino with the extraction algorithm used to convert its output into ILL$_{\li, \Diamond, \Box}$ derivations~\cite{kogkalidis-moortgat-moot:2020:LREC}; 
together, the two faithfully replicate the data generating process our system has been trained on, modulo the manual correction phase of~\citet{van2013large}.
We query Alpino for the globally optimal parse of each sample in the test set (enforcing no time constraints), perform the conversion and log the results in Table~\ref{tab:numbers}.

Our model achieves remarkable performance even in the greedy setting, especially considering the rigidity of our metrics.
Untyped term accuracy conveys the percentage of sentences for which the function-argument structure has been perfectly captured.
Typed term accuracy is even stricter; the added requirement of a correct proof frame practically translates to no erroneous assignments of part-of-speech and syntactic phrase tags or dependency labels.
Keeping in mind that dependency information are already incorporated in the proof frame, obtaining the correct proof structure fully subsumes dependency parsing.

The filtering criteria of the previous paragraph yield significant benefits when combined with beam search, allowing us to circumvent logically unsound analyses regardless of their sequence scores.
It is worth noting that our metrics place the model's bottleneck at the supertagging rather than the permutation component.
Term accuracy closely follows along (and actually surpasses, in the untyped case) frame accuracy.
This is further evidenced when providing the ground truth types as input to the parser, in which case term accuracy reaches as high as $85.4\%$, indicative of the high expressive power of Sinkhorn on top of the the bi-modal encoder's contextualization.
On the negative side, the strong reliance on correct type assignments means that a single mislabeled word can heavily skew the parse outcome, but also hints at increasing returns from improvements in the decoding architecture.

\section{Related Work}
\label{sec:lit}
Our work bears semblances to other neural methodologies related to syntactic/semantic parsing.
Sequence-to-sequence models have been successfully employed in the past to decode directly into flattened representations of parse trees~\cite{wiseman2016sequence, buys2017robust, li2018seq2seq}.
In dependency parsing literature, head selection involves building word representations that act as classifying functions over other words~\cite{zhang2017dependency}, similar to our dot-product weighting between atoms.

Akin to graph-based parsers~\cite{ji2019graph, zhang-etal-2019-amr}, our model generates parse structures in the form of graphs.
In our case, however, graph nodes correspond to syntactic primitives (atomic types \& dependencies) rather than words, while the discovery of the graph structure is subject to hard constraints imposed by the decoder's output.

Transcription to formal expressions (logical forms, $\lambda$-terms, database queries and executable program instructions) has also been a prominent theme in NLP literature, using statistical methods~\cite{zettlemoyer2012learning} or structurally-constrained decoders~\cite{dong2016language, xiao2016sequence, liu2018discourse, cheng2019learning}.
Unlike prior approaches, the decoding we employ here is unhindered by explicit structure; instead, parsing is handled in parallel across the entire sequence by the Sinkhorn operator, which biases the output towards structural correctness while requiring neither backtracking nor iterative processing.
More importantly, the $\lambda$-terms we generate are not in themselves the product of a neural decoding process, but rather a corollary of the isomorphic relation between ILL$_\li$ proofs and linear $\lambda$-calculus programs.

In machine learning literature, Sinkhorn-based networks have been gaining popularity as a means of learning latent permutations of visual or synthetic data~\cite{mena2018learning} or imposing permutation invariance for set-theoretic learning~\cite{grover2018stochastic}, with so far limited adoption in the linguistic setting~\cite{tay2020sparse, swanson2020rationalizing}.
In contrast to prior applications of Sinkhorn as a final classification layer, we use it over chain element representations that have been mutually contextualized, rather than set elements vectorized in isolation.
Our benchmarks, combined with the assignment-preserving property of the operator, hint towards potential benefits from adopting it in a similar fashion across other parsing tasks.

\section{Conclusion}
\label{sec:conc}
We have introduced neural proof nets, a data-driven perspective on the proof nets of ILL$_\li$, and successfully employed them on the demanding task of transcribing raw text to proofs and computational terms of the linear $\lambda$-calculus.
The terms construed constitute type-safe abstract program skeletons that are free to interpret within arbitrary domains, fulfilling the role of a practical intermediary between text and meaning.
Used as-is, they can find direct application in logic-driven models of natural language inference~\cite{abzianidze-2016-natural}.

Our architecture marks a departure from other parsing approaches, owing to the novel use of the Sinkhorn operator, which renders it both fully parallel and backtrack-free, but also logically grounded.
It is general enough to apply to a variety of grammar formalisms inheriting from linear logic; if augmented with Gumbel sampling~\cite{mena2018learning}, it can further a provide a probabilistic means to account for derivational ambiguity.
Viewed as a means of exposing deep tecto-grammatic structure, it paves the way for graph-theoretic approaches at syntax-aware sentential meaning representations.

\section*{Acknowledgements}
We would like to thank the anonymous reviewers for their detailed feedback, which helped improve the presentation of the paper.
Konstantinos and Michael are supported by the Dutch Research Council (NWO) under the scope of the project ``A composition calculus for vector-based semantic modelling with a localization for Dutch'' (360-89-070).

\bibliography{main}

\begin{thebibliography}{57}
\expandafter\ifx\csname natexlab\endcsname\relax\def\natexlab#1{#1}\fi

\bibitem[{Abzianidze(2016)}]{abzianidze-2016-natural}
Lasha Abzianidze. 2016.
\newblock \href {https://doi.org/10.18653/v1/S16-2007} {Natural solution to
  {F}ra{C}a{S} entailment problems}.
\newblock In \emph{Proceedings of the Fifth Joint Conference on Lexical and
  Computational Semantics}, pages 64--74, Berlin, Germany. Association for
  Computational Linguistics.

\bibitem[{Ba et~al.(2016)Ba, Kiros, and Hinton}]{ba2016layer}
Jimmy~Lei Ba, Jamie~Ryan Kiros, and Geoffrey~E Hinton. 2016.
\newblock Layer normalization.
\newblock \emph{arXiv preprint arXiv:1607.06450v1}.

\bibitem[{Bangalore and Joshi(1999)}]{bangalore1999supertagging}
Srinivas Bangalore and Aravind~K Joshi. 1999.
\newblock Supertagging: An approach to almost parsing.
\newblock \emph{Computational linguistics}, 25(2):237--265.

\bibitem[{Bhargava and Penn(2020)}]{bhargava-penn-2020-supertagging}
Aditya Bhargava and Gerald Penn. 2020.
\newblock \href {https://www.aclweb.org/anthology/2020.repl4nlp-1.23}
  {Supertagging with {CCG} primitives}.
\newblock In \emph{Proceedings of the 5th Workshop on Representation Learning
  for NLP}, pages 194--204, Online. Association for Computational Linguistics.

\bibitem[{Bouma and van Noord(2017)}]{bouma-van-noord-2017-increasing}
Gosse Bouma and Gertjan van Noord. 2017.
\newblock \href {https://www.aclweb.org/anthology/W17-0403} {Increasing return
  on annotation investment: The automatic construction of a {U}niversal
  {D}ependency treebank for {D}utch}.
\newblock In \emph{Proceedings of the {N}o{D}a{L}i{D}a 2017 Workshop on
  Universal Dependencies ({UDW} 2017)}, pages 19--26, Gothenburg, Sweden.
  Association for Computational Linguistics.

\bibitem[{Bouma et~al.(2001)Bouma, van Noord, and Malouf}]{bouma2001alpino}
Gosse Bouma, Gertjan van Noord, and Robert Malouf. 2001.
\newblock Alpino: Wide-coverage computational analysis of dutch.
\newblock In \emph{Computational linguistics in the Netherlands 2000}, pages
  45--59. Brill Rodopi.

\bibitem[{de~Bruijn(1979)}]{de1979wiskundigen}
Nicolaas~Govert de~Bruijn. 1979.
\newblock Wiskundigen, let op uw {N}ederlands.
\newblock \emph{Euclides}, 55(juni/juli):429--435.

\bibitem[{Buys and Blunsom(2017)}]{buys2017robust}
Jan Buys and Phil Blunsom. 2017.
\newblock Robust incremental neural semantic graph parsing.
\newblock In \emph{Proceedings of the 55th Annual Meeting of the Association
  for Computational Linguistics (Volume 1: Long Papers)}, pages 1215--1226.

\bibitem[{Che et~al.(2018)Che, Liu, Wang, Zheng, and
  Liu}]{che-etal-2018-towards}
Wanxiang Che, Yijia Liu, Yuxuan Wang, Bo~Zheng, and Ting Liu. 2018.
\newblock \href {https://doi.org/10.18653/v1/K18-2005} {Towards better {UD}
  parsing: Deep contextualized word embeddings, ensemble, and treebank
  concatenation}.
\newblock In \emph{Proceedings of the {C}o{NLL} 2018 Shared Task: Multilingual
  Parsing from Raw Text to Universal Dependencies}, pages 55--64, Brussels,
  Belgium. Association for Computational Linguistics.

\bibitem[{Cheng et~al.(2019)Cheng, Reddy, Saraswat, and
  Lapata}]{cheng2019learning}
Jianpeng Cheng, Siva Reddy, Vijay Saraswat, and Mirella Lapata. 2019.
\newblock Learning an executable neural semantic parser.
\newblock \emph{Computational Linguistics}, 45(1):59--94.

\bibitem[{Danos and Regnier(1989)}]{multiplicatives}
Vincent Danos and Laurent Regnier. 1989.
\newblock The structure of multiplicatives.
\newblock \emph{Archive for Mathematical Logic}, 28:181--203.

\bibitem[{Devlin et~al.(2019)Devlin, Chang, Lee, and
  Toutanova}]{devlin2018bert}
Jacob Devlin, Ming-Wei Chang, Kenton Lee, and Kristina Toutanova. 2019.
\newblock Bert: Pre-training of deep bidirectional transformers for language
  understanding.
\newblock In \emph{Proceedings of the 2019 Conference of the North American
  Chapter of the Association for Computational Linguistics: Human Language
  Technologies, Volume 1 (Long and Short Papers)}, pages 4171--4186.

\bibitem[{Dong and Lapata(2016)}]{dong2016language}
Li~Dong and Mirella Lapata. 2016.
\newblock Language to logical form with neural attention.
\newblock In \emph{Proceedings of the 54th Annual Meeting of the Association
  for Computational Linguistics (Volume 1: Long Papers)}, pages 33--43.

\bibitem[{Girard(1987)}]{girard1987linear}
Jean-Yves Girard. 1987.
\newblock Linear logic.
\newblock \emph{Theoretical computer science}, 50(1):1--101.

\bibitem[{Girard et~al.(1988)Girard, Lafont, and Taylor}]{glt}
Jean-Yves Girard, Yves Lafont, and P.~Taylor. 1988.
\newblock \emph{Proofs and Types}.
\newblock Cambridge Tracts in Theoretical Computer Science 7. Cambridge
  University Press.

\bibitem[{de~Groote(2001)}]{de2001towards}
Philippe de~Groote. 2001.
\newblock Towards abstract categorial grammars.
\newblock In \emph{Proceedings of the 39th Annual Meeting of the Association
  for Computational Linguistics}, pages 252--259.

\bibitem[{de~Groote and Retor{\'e}(1996)}]{degrooteRetore}
Philippe de~Groote and Christian Retor{\'e}. 1996.
\newblock \href {https://hal.archives-ouvertes.fr/hal-00823554} {{On the
  semantic readings of proof-nets}}.
\newblock In \emph{{Proceedings Formal grammar}}, pages 57--70, Prague, Czech
  Republic. {FoLLI}.

\bibitem[{Grover et~al.(2019)Grover, Wang, Zweig, and
  Ermon}]{grover2018stochastic}
Aditya Grover, Eric Wang, Aaron Zweig, and Stefano Ermon. 2019.
\newblock \href {https://openreview.net/forum?id=H1eSS3CcKX} {Stochastic
  optimization of sorting networks via continuous relaxations}.
\newblock In \emph{International Conference on Learning Representations}.

\bibitem[{Guerrini(1999)}]{pnlinear}
Stefano Guerrini. 1999.
\newblock Correctness of multiplicative proof nets is linear.
\newblock In \emph{Fourteenth Annual {IEEE} Symposium on Logic in Computer
  Science}, pages 454--263. {IEEE} Computer Science Society.

\bibitem[{Hendrycks and Gimpel(2016)}]{hendrycks2016bridging}
Dan Hendrycks and Kevin Gimpel. 2016.
\newblock Bridging nonlinearities and stochastic regularizers with gaussian
  error linear units.

\bibitem[{Ji et~al.(2019)Ji, Wu, and Lan}]{ji2019graph}
Tao Ji, Yuanbin Wu, and Man Lan. 2019.
\newblock Graph-based dependency parsing with graph neural networks.
\newblock In \emph{Proceedings of the 57th Annual Meeting of the Association
  for Computational Linguistics}, pages 2475--2485.

\bibitem[{Kanovich(1994)}]{Kanovich94}
Max~I. Kanovich. 1994.
\newblock The complexity of horn fragments of linear logic.
\newblock \emph{Annals of Pure and Applied Logic}, 69(2-3):195--241.

\bibitem[{Kogkalidis et~al.(2019)Kogkalidis, Moortgat, and
  Deoskar}]{kogkalidis2019constructive}
Konstantinos Kogkalidis, Michael Moortgat, and Tejaswini Deoskar. 2019.
\newblock Constructive type-logical supertagging with self-attention networks.
\newblock In \emph{Proceedings of the 4th Workshop on Representation Learning
  for NLP (RepL4NLP-2019)}, pages 113--123.

\bibitem[{Kogkalidis et~al.(2020)Kogkalidis, Moortgat, and
  Moot}]{kogkalidis-moortgat-moot:2020:LREC}
Konstantinos Kogkalidis, Michael Moortgat, and Richard Moot. 2020.
\newblock \href {https://www.aclweb.org/anthology/2020.lrec-1.647} {Æthel:
  Automatically extracted typelogical derivations for dutch}.
\newblock In \emph{Proceedings of The 12th Language Resources and Evaluation
  Conference}, pages 5259--5268, Marseille, France. European Language Resources
  Association.

\bibitem[{Kubota and Levine(2020)}]{KubotaLevine20}
Ysuke Kubota and Robert Levine. 2020.
\newblock \emph{Type-Logical Syntax}.
\newblock MIT Press.

\bibitem[{Lamarche(2008)}]{lamarcheEssential}
Fran{\c c}ois Lamarche. 2008.
\newblock \href {https://hal.inria.fr/inria-00347336/file/prfnet1.pdf} {Proof
  nets for intuitionistic linear logic: Essential nets}.
\newblock Research report, INRIA Nancy.

\bibitem[{Lambek(1958)}]{lambek1958mathematics}
Joachim Lambek. 1958.
\newblock The mathematics of sentence structure.
\newblock \emph{The American Mathematical Monthly}, 65(3):154--170.

\bibitem[{Li et~al.(2018)Li, Cai, He, and Zhao}]{li2018seq2seq}
Zuchao Li, Jiaxun Cai, Shexia He, and Hai Zhao. 2018.
\newblock Seq2seq dependency parsing.
\newblock In \emph{Proceedings of the 27th International Conference on
  Computational Linguistics}, pages 3203--3214.

\bibitem[{Lincoln(1995)}]{lincoln}
Patrick Lincoln. 1995.
\newblock Deciding provability of linear logic formulas.
\newblock In Jean-Yves Girard, Yves Lafont, and Laurent Regnier, editors,
  \emph{Advances in Linear Logic}, pages 109--122. Cambridge University Press.

\bibitem[{Liu et~al.(2018)Liu, Cohen, and Lapata}]{liu2018discourse}
Jiangming Liu, Shay~B Cohen, and Mirella Lapata. 2018.
\newblock Discourse representation structure parsing.
\newblock In \emph{Proceedings of the 56th Annual Meeting of the Association
  for Computational Linguistics (Volume 1: Long Papers)}, pages 429--439.

\bibitem[{Loshchilov and Hutter(2018)}]{loshchilov2018fixing}
Ilya Loshchilov and Frank Hutter. 2018.
\newblock Fixing weight decay regularization in adam.

\bibitem[{Mena et~al.(2018)Mena, Belanger, Linderman, and
  Snoek}]{mena2018learning}
Gonzalo Mena, David Belanger, Scott Linderman, and Jasper Snoek. 2018.
\newblock \href {https://openreview.net/forum?id=Byt3oJ-0W} {Learning latent
  permutations with {G}umbel-{S}inkhorn networks}.
\newblock In \emph{International Conference on Learning Representations}.

\bibitem[{Moortgat(1996)}]{Moortgat96}
Michael Moortgat. 1996.
\newblock Multimodal linguistic inference.
\newblock \emph{Journal of Logic, Language and Information}, 5(3/4):349--385.

\bibitem[{Morrill(2014)}]{Morrill14}
Glyn Morrill. 2014.
\newblock A categorial type logic.
\newblock In \emph{Categories and Types in Logic, Language, and Physics -
  Essays Dedicated to Jim Lambek on the Occasion of His 90th Birthday}, volume
  8222 of \emph{Lecture Notes in Computer Science}, pages 331--352. Springer.

\bibitem[{M{\"u}ller et~al.(2019)M{\"u}ller, Kornblith, and
  Hinton}]{muller2019does}
Rafael M{\"u}ller, Simon Kornblith, and Geoffrey~E Hinton. 2019.
\newblock When does label smoothing help?
\newblock In \emph{Advances in Neural Information Processing Systems}, pages
  4696--4705.

\bibitem[{Murawski and Ong(2000)}]{murong}
Andrzej~S. Murawski and C.-H.~Luke Ong. 2000.
\newblock Dominator trees and fast verification of proof nets.
\newblock In \emph{Logic in Computer Science}, pages 181--191.

\bibitem[{Muskens(2001)}]{muskens2001lambda}
Reinhard Muskens. 2001.
\newblock Lambda grammars and the syntax-semantics interface.
\newblock In \emph{Proceedings of the 13th Amsterdam Colloquium}, pages
  150--155.

\bibitem[{Muskens and Sadrzadeh(2018)}]{MuskensSadrzadeh18}
Reinhard Muskens and Mehrnoosh Sadrzadeh. 2018.
\newblock Static and dynamic vector semantics for lambda calculus models of
  natural language.
\newblock \emph{Journal of Language Modelling}, 6(2):319--351.

\bibitem[{van Noord et~al.(2013)van Noord, Bouma, van Eynde, de~Kok, van~der
  Linde, Schuurman, Sang, and Vandeghinste}]{van2013large}
Gertjan van Noord, Gosse Bouma, Frank van Eynde, Daniel de~Kok, Jelmer van~der
  Linde, Ineke Schuurman, Erik Tjong~Kim Sang, and Vincent Vandeghinste. 2013.
\newblock Large scale syntactic annotation of written dutch: Lassy.
\newblock In \emph{Essential speech and language technology for Dutch}, pages
  147--164. Springer, Berlin, Heidelberg.

\bibitem[{Press and Wolf(2017)}]{press2017using}
Ofir Press and Lior Wolf. 2017.
\newblock Using the output embedding to improve language models.
\newblock In \emph{Proceedings of the 15th Conference of the European Chapter
  of the Association for Computational Linguistics: Volume 2, Short Papers},
  pages 157--163.

\bibitem[{Roorda(1991)}]{roorda}
Dirk Roorda. 1991.
\newblock \emph{Resource Logics: Proof-theoretical Investigations}.
\newblock Ph.D. thesis, Universiteit van Amsterdam.

\bibitem[{Sinkhorn(1964)}]{sinkhorn1964relationship}
Richard Sinkhorn. 1964.
\newblock A relationship between arbitrary positive matrices and doubly
  stochastic matrices.
\newblock \emph{The annals of mathematical statistics}, 35(2):876--879.

\bibitem[{S{\o}rensen and Urzyczyn(2006)}]{sorensen2006lectures}
Morten~Heine S{\o}rensen and Pawel Urzyczyn. 2006.
\newblock \emph{Lectures on the Curry-Howard isomorphism}.
\newblock Elsevier.

\bibitem[{Swanson et~al.(2020)Swanson, Yu, and Lei}]{swanson2020rationalizing}
Kyle Swanson, Lili Yu, and Tao Lei. 2020.
\newblock Rationalizing text matching: Learning sparse alignments via optimal
  transport.
\newblock \emph{arXiv preprint arXiv:2005.13111}.

\bibitem[{Tay et~al.(2020)Tay, Bahri, Yang, Metzler, and Juan}]{tay2020sparse}
Yi~Tay, Dara Bahri, Liu Yang, Donald Metzler, and Da-Cheng Juan. 2020.
\newblock Sparse sinkhorn attention.
\newblock \emph{arXiv preprint arXiv:2002.11296v1}.

\bibitem[{Troelstra and Schwichtenberg(2000)}]{bpt}
Anne~Sjerp Troelstra and Helmut Schwichtenberg. 2000.
\newblock \emph{Basic Proof Theory}, 2 edition, volume~43 of \emph{Cambridge
  Tracts in Theoretical Computer Science}.
\newblock Cambridge University Press.

\bibitem[{Vaswani et~al.(2016)Vaswani, Bisk, Sagae, and
  Musa}]{vaswani2016supertagging}
Ashish Vaswani, Yonatan Bisk, Kenji Sagae, and Ryan Musa. 2016.
\newblock Supertagging with lstms.
\newblock In \emph{Proceedings of the 2016 Conference of the North American
  Chapter of the Association for Computational Linguistics: Human Language
  Technologies}, pages 232--237.

\bibitem[{Vaswani et~al.(2017)Vaswani, Shazeer, Parmar, Uszkoreit, Jones,
  Gomez, Kaiser, and Polosukhin}]{vaswani2017attention}
Ashish Vaswani, Noam Shazeer, Niki Parmar, Jakob Uszkoreit, Llion Jones,
  Aidan~N Gomez, {\L}ukasz Kaiser, and Illia Polosukhin. 2017.
\newblock Attention is all you need.
\newblock In \emph{Advances in neural information processing systems}, pages
  5998--6008.

\bibitem[{de~Vries et~al.(2019)de~Vries, van Cranenburgh, Bisazza, Caselli, van
  Noord, and Nissim}]{de2019bertje}
Wietse de~Vries, Andreas van Cranenburgh, Arianna Bisazza, Tommaso Caselli,
  Gertjan van Noord, and Malvina Nissim. 2019.
\newblock {BERT}je: A {D}utch {BERT} model.
\newblock \emph{arXiv preprint arXiv:1912.09582v1}.

\bibitem[{Wadler(1993)}]{wadler1993taste}
Philip Wadler. 1993.
\newblock A taste of linear logic.
\newblock In \emph{International Symposium on Mathematical Foundations of
  Computer Science}, pages 185--210. Springer.

\bibitem[{Wang et~al.(2020)Wang, Zhao, Lioma, Li, Zhang, and
  Simonsen}]{Wang2020Encoding}
Benyou Wang, Donghao Zhao, Christina Lioma, Qiuchi Li, Peng Zhang, and
  Jakob~Grue Simonsen. 2020.
\newblock \href {https://openreview.net/forum?id=Hke-WTVtwr} {Encoding word
  order in complex embeddings}.
\newblock In \emph{International Conference on Learning Representations}.

\bibitem[{Wiseman and Rush(2016)}]{wiseman2016sequence}
Sam Wiseman and Alexander~M Rush. 2016.
\newblock Sequence-to-sequence learning as beam-search optimization.
\newblock In \emph{Proceedings of the 2016 Conference on Empirical Methods in
  Natural Language Processing}, pages 1296--1306.

\bibitem[{Xiao et~al.(2016)Xiao, Dymetman, and Gardent}]{xiao2016sequence}
Chunyang Xiao, Marc Dymetman, and Claire Gardent. 2016.
\newblock Sequence-based structured prediction for semantic parsing.
\newblock In \emph{Proceedings of the 54th Annual Meeting of the Association
  for Computational Linguistics (Volume 1: Long Papers)}, pages 1341--1350.

\bibitem[{Xu et~al.(2015)Xu, Auli, and Clark}]{xu2015ccg}
Wenduan Xu, Michael Auli, and Stephen Clark. 2015.
\newblock Ccg supertagging with a recurrent neural network.
\newblock In \emph{Proceedings of the 53rd Annual Meeting of the Association
  for Computational Linguistics and the 7th International Joint Conference on
  Natural Language Processing (Volume 2: Short Papers)}, pages 250--255.

\bibitem[{Zettlemoyer and Collins(2012)}]{zettlemoyer2012learning}
Luke~S Zettlemoyer and Michael Collins. 2012.
\newblock Learning to map sentences to logical form: Structured classification
  with probabilistic categorial grammars.
\newblock \emph{arXiv preprint arXiv:1207.1420v1}.

\bibitem[{Zhang et~al.(2019)Zhang, Ma, Duh, and
  Van~Durme}]{zhang-etal-2019-amr}
Sheng Zhang, Xutai Ma, Kevin Duh, and Benjamin Van~Durme. 2019.
\newblock \href {https://doi.org/10.18653/v1/P19-1009} {{AMR} parsing as
  sequence-to-graph transduction}.
\newblock In \emph{Proceedings of the 57th Annual Meeting of the Association
  for Computational Linguistics}, pages 80--94, Florence, Italy. Association
  for Computational Linguistics.

\bibitem[{Zhang et~al.(2017)Zhang, Cheng, and Lapata}]{zhang2017dependency}
Xingxing Zhang, Jianpeng Cheng, and Mirella Lapata. 2017.
\newblock Dependency parsing as head selection.
\newblock In \emph{Proceedings of the 15th Conference of the European Chapter
  of the Association for Computational Linguistics: Volume 1, Long Papers},
  pages 665--676.

\end{thebibliography}
\bibliographystyle{acl_natbib}
\nocite{roorda, lamarcheEssential, degrooteRetore, ba2016layer, hendrycks2016bridging}

\clearpage

\appendix
\section{Appendix}
\label{sec:appendix}
\subsection{Model}
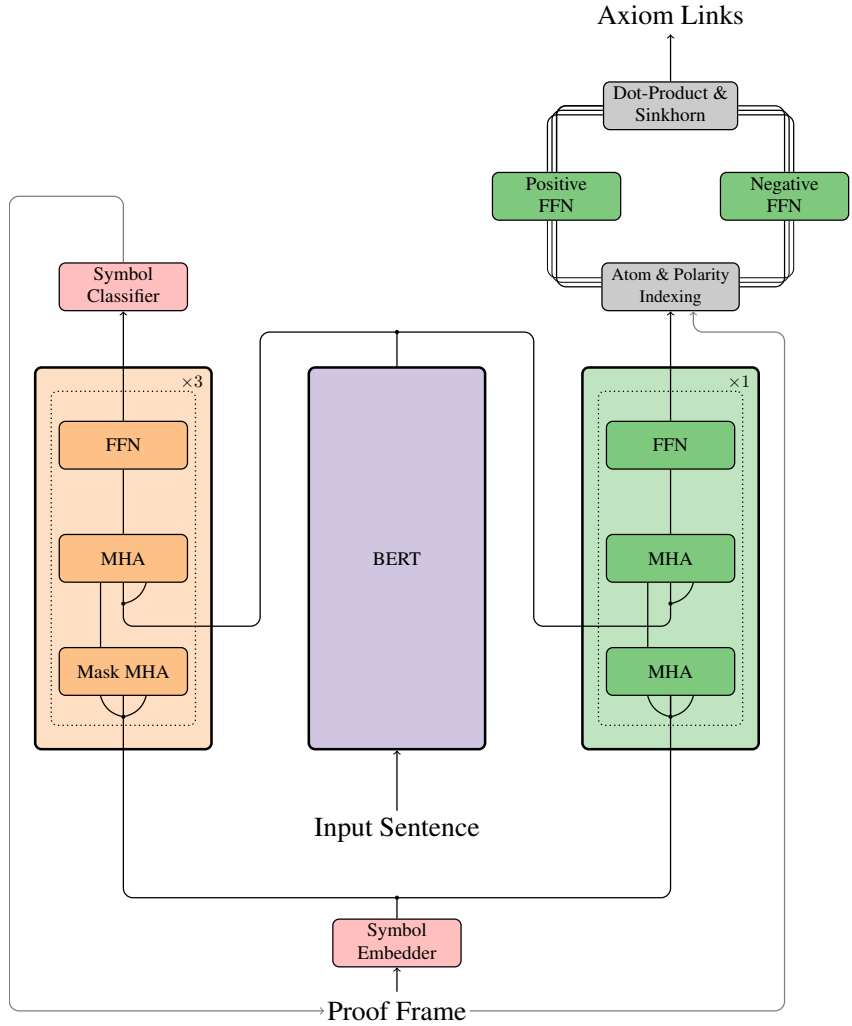
\begin{figure*}
    \centering
     \scalebox{0.6}{


\definecolor{first}{RGB}{82,82,82}
\definecolor{enc}{RGB}{253,192,134}
\definecolor{dec}{RGB}{127,201,127}
\definecolor{dec2}{RGB}{140,220,140}
\definecolor{emb}{RGB}{190,174,212}


\begin{tikzpicture}[every text node part/.style={align=center},
 every node/.style={transform shape},
 =>stealth,
 scale=1,
 block/.style={rectangle, inner sep=0pt, minimum width=110pt, minimum height=40pt, rounded corners, ultra thick},
 shortblock/.style={rectangle, inner sep=0pt, minimum width=80pt, minimum height=30pt, rounded corners, thick},
 tallblock/.style={rectangle, inner sep=0pt, minimum width=110pt, minimum height=240pt, rounded corners, ultra thick},
 dotblock/.style={rectangle, inner sep=0pt, minimum width=90pt, minimum height=210pt, rounded corners, thick, dotted, draw=black},
 arrow/.style={->, thick},
 tharrow/.style={->, thick},
 str/.style={rectangle, inner sep=0pt, minimum width=80pt, minimum height=20pt, outer sep=2pt},
 dot/.style={circle, inner sep=0pt, minimum size=2pt, fill=black, draw=black}
]
    \clip (-2.5, -5) rectangle (16,18.5);
    \node[str] (sentence) at (6, 0) {\LARGE Input Sentence};
    \node[str] (frame) at (6, -4) {\LARGE Proof Frame};
    \node[str] (out) at (12, 18) {\LARGE Axiom Links};
    
    \node[tallblock, draw=black, fill=emb!70] (bert) at (6, 6) {\large BERT};
    \node[tallblock, draw=black, fill=enc!50] (super) at (0, 6) {};
    \node[tallblock, draw=black, fill=dec!50] (super) at (12, 6) {};
    \node[dotblock] (cross1) at (0, 6) {};
    \node[str] (cross1t) at (1.5, 9.9) {$\times 3$};
    \node[shortblock, draw=black, fill=enc] (satn) at (0,3.5) {\large Mask MHA};
    \node[shortblock, draw=black, fill=enc] (catn) at (0,6) {\large MHA};
    \node[shortblock, draw=black, fill=enc] (ffn) at (0,8.5) {\large FFN};
    \node[dotblock] (cross2) at (12, 6) {};
    \node[str] (cross2t) at (13.5, 9.9) {$\times 1$};
    \node[shortblock, fill=pink, draw=black] (emb) at (6, -2.5) {\large \begin{tabular}{c}Symbol\\Embedder\end{tabular}};
    \node[dot] (embout) at (6, -1.5) {};
    \node[shortblock, fill=pink, draw=black] (class) at (0, 12) {\large \begin{tabular}{c}Symbol\\Classifier\end{tabular}};
    \node[shortblock, fill=first!30, draw=black] (index) at (12, 12)
    {\begin{tabular}{c}Atom \& Polarity\\Indexing\end{tabular}};
    \node[shortblock, fill=dec, draw=black] (pffn) at (9.5, 14)
    {\large \begin{tabular}{c}Positive\\FFN\end{tabular}};
    \node[shortblock, fill=dec, draw=black] (nffn) at (14.5, 14)
    {\large \begin{tabular}{c}Negative\\FFN\end{tabular}};
    \draw[thick, rounded corners=6pt] (index) -- (9.5, 12) -- (pffn);
    \draw[thick, rounded corners=6pt] (index) -- (14.5, 12) -- (nffn);
    \draw[thin, rounded corners=6pt] ($(index.east) + (0, 0.1)$) -- (14.6, 12.1) -- ($(nffn.south) + (0.1, 0)$);
    \draw[thin, rounded corners=6pt] ($(index.east) + (0, 0.2)$) -- (14.7, 12.2) -- ($(nffn.south) + (0.2, 0)$);
    \draw[thin, rounded corners=6pt] ($(index.west) + (0, 0.1)$) -- (9.4, 12.1) -- ($(pffn.south) - (0.1, 0)$);
    \draw[thin, rounded corners=6pt] ($(index.west) + (0, 0.2)$) -- (9.3, 12.2) -- ($(pffn.south) - (0.2, 0)$);
    \node[shortblock, fill=first!30, draw=black] (dp) at (12, 16)    {\large \begin{tabular}{c}Dot-Product \&\\Sinkhorn\end{tabular}};
    \draw[thick, rounded corners=6pt] (pffn) -- (9.5, 16) -- (dp);
    \draw[thick, rounded corners=6pt] ($(pffn.north)-(0.1, 0)$) -- (9.4, 15.9) -- ($(dp.west) - (0, 0.1)$);
    \draw[thick, rounded corners=6pt] ($(pffn.north)-(0.2, 0)$) -- (9.3, 15.8) -- ($(dp.west) - (0, 0.2)$);
    \draw[thick, rounded corners=6pt] (pffn) -- (9.5, 16) -- (dp);
    \draw[thick, rounded corners=6pt] (nffn) -- (14.5, 16) -- (dp);
    \draw[thick, rounded corners=6pt] ($(nffn.north)+(0.1, 0)$) -- (14.6, 15.9) -- ($(dp.east) - (0, 0.1)$);
    \draw[thick, rounded corners=6pt] ($(nffn.north)+(0.2, 0)$) -- (14.7, 15.8) -- ($(dp.east) - (0, 0.2)$);
    \draw[tharrow] (dp) edge (out);

    \draw[tharrow] (sentence) edge (bert);
    
    \node[dot] (lcatnin) at (12, 2.5) {};
    \node[dot] (ssatn) at (0, 2.5) {};
    \node[dot] (catnin) at (0, 5) {};
    \node[dot] (lsatnin) at (12, 5) {};
    \node[dot] (bertout) at (6, 11) {};

    \draw[thick] (bert) edge (bertout);
    
    \draw[thick] (ssatn) edge (satn);
    \draw[thick] (ssatn) edge[bend left] ($(satn.south) - (0.5, 0)$);
    \draw[thick] (ssatn) edge[bend right] ($(satn.south) + (0.5, 0)$);
    \draw[thick] ($(satn.north) - (0.5, 0)$) edge ($(catn.south) - (0.5, 0)$);
    \draw[thick] ($(catn.north)$) edge (ffn);
    \draw[thick] (catnin) edge (catn);
    \draw[thick] (catnin) edge[bend right] ($(catn.south) + (0.5, 0)$);
    \draw[thick, rounded corners=6pt] (bertout) -- (3, 11) -- (3, 4.5) -- (0, 4.5) -- (catnin);
    \draw[tharrow] (ffn) edge (class);
    
    \node[shortblock, draw=black, fill=dec] (lcatn) at (12, 3.5) {\large MHA};
    \node[shortblock, draw=black, fill=dec] (lsatn) at (12, 6) {\large MHA};
    \node[shortblock, draw=black, fill=dec] (lffn) at (12, 8.5) {\large FFN};
    
    \draw[thick] (lsatnin) edge[bend right] ($(lsatn.south) + (0.5, 0)$);
    \draw[thick] (lsatnin) edge (lsatn);
    \draw[thick, rounded corners=6pt] (bertout) -- (9, 11) -- (9, 4.5) -- (12, 4.5) -- (lsatnin);
    \draw[thick] (lcatnin) edge (lcatn);
    \draw[thick] (lcatnin) edge[bend right] ($(lcatn.south) + (0.5, 0)$);
    \draw[thick] (lcatnin) edge[bend left] ($(lcatn.south) - (0.5, 0)$);
    \draw[thick] ($(lcatn.north) - (0.5, 0)$) edge ($(lsatn.south) - (0.5, 0)$);
    \draw[thick] (lsatn) edge (lffn);
    \draw[tharrow] (lffn) edge (index);
    
    \draw[thick] (emb) edge (embout);
    \draw[tharrow] (frame) edge (emb);
    \draw[thick, rounded corners=6pt] (embout) -- (0, -1.5) -- (ssatn);
    \draw[thick, rounded corners=6pt] (embout) -- (12, -1.5) -- (lcatn);
    \draw[tharrow, color=gray, rounded corners=6pt] (class) -- (0, 14) -- (-2.5, 14) -- (-2.5, -4) -- ($(frame.west)$);

    \draw[tharrow, rounded corners=6pt, color=gray] (frame) -- (14.5, -4) -- (14.5, 11) -- (12.5, 11) -- ($(index.south) + (0.5, 0)$);
\end{tikzpicture}

}  

    \caption{Schematic diagram of the full network architecture. The supertagger (orange, left) iteratively generates a proof frame by attending over the currently available part of it plus the full input sentence. The axiom linker (green, right) contextualizes the complete proof frame by attending over it as well as the sentence. Representations of atomic formulas are gathered and transformed according to their polarity, and their Sinkhorn-activated dot-product attention is computed. Discretization of the result yields a permutation matrix denoting axiom links for each unique atomic type in the proof frame. The final output is a proof structure, i.e. the pair of a proof frame and its axiom links.}
    \label{fig:network}
\end{figure*}

Table~\ref{tab:model} presents model hyper-parameters, as selected by greedy grid search.
An illustration of the model can be seen in Figure~\ref{fig:network}.

\begin{table}[h]
    \centering
    \newcolumntype{L}{>{\raggedright\arraybackslash}p{0.7\columnwidth}}
    \begin{tabularx}{0.99\columnwidth}{@{}Lc@{}}
    \textbf{Parameter} & \textbf{Value} \\
    \midrule
    \multicolumn{2}{c}{BERTje (\textit{BERT-Base})} \\
    \midrule
    \# Layers & $12$ \\
    \# Self-attention heads & $12$ \\
    Feed-forward dimensionality & $3\,072$ \\ 
    Feed-forward activation & GELU \\
    Input/output dimensionality & $768$ \\
    Vocabulary size & $30\,000$ \\
    \midrule
    \multicolumn{2}{c}{Decoder} \\
    \midrule
    \# Layers & $3$ \\ 
    \# Self-attention heads & $8$ \\ 
    \# Encoder-attention heads & $8$ \\
    Feed-forward dimensionality & $512$ \\ 
    Input/output dimensionality & $256$ \\ 
    Vocabulary size & 58 \\
    \midrule
    \multicolumn{2}{c}{Bi-modal Encoder} \\
    \midrule
    \# Layers & $1$ \\ 
    \# Self-attention heads & $8$ \\ 
    \# Encoder-attention heads & $8$ \\ 
    Feed-forward dimensionality & $512$ \\ 
    Feed-forward activation & GELU \\
    Input/output dimensionality & $256$ \\
    \midrule
    \multicolumn{2}{c}{Pre-Sinkhorn Transformations} \\
    \midrule
    Input/Feed-forward dimensionality & $256$ \\ 
    Feed-forward activation & GELU \\ 
    Output dimensionality & $32$ \\
    Output activation & LayerNorm
    \end{tabularx}
    \caption{Model hyper-parameters}
    \label{tab:model}
\end{table}%

\subsection{Optimization}
We train with an adaptive learning rate following~\citet{vaswani2017attention}, such that the learning rate at optimization step $i$ is given as:
\[
    768^{-0.5} \cdot \mathrm{min}\left(i^{-0.5}, \ i \cdot \mathrm{warmup\_steps}^{-1.5}\right)
\]
For BERT parameters, learning rate is scaled by $0.1$.
We freeze the oversized word embedding layer to reduce training costs and avoid overfitting.
Optimization hyper-parameters are presented in Table~\ref{tab:optim}.

We provide strict teacher guidance when learning axiom links, whereby the network is provided with the original proof frame symbol sequence instead of the predicted one.
To speed up computation, positive and negative indexes are arranged per-length rather than type for each batch; this allows us to process symbol transformations, dot-product attentions and Sinkhorn activations in parallel for many types across many sentences.
During training, we set the number of Sinkhorn iterations to $5$; lower values are more difficult to reach convergence with, hurting performance, whereas higher values can easily lead to vanishing gradients, impeding learning~\cite{grover2018stochastic}.

\begin{table}[h]
    \centering
    \newcolumntype{L}{>{\raggedright\arraybackslash}p{0.7\columnwidth}}
    \begin{tabularx}{0.99\columnwidth}{@{}Lc@{}}
    \textbf{Parameter} & \textbf{Value} \\
    \midrule
    Batch size & $32$ \\
    Warmup epochs & $5$ \\
    Weight decay & $10^{-5}$ \\
    Weight decay (BERT) & $0$ \\
    LR scale (BERT) & $0.1$ \\ 
    LR scale (BERT embedding ) & $0$ \\
    Dropout rate & $0.1$ \\ 
    Label smoothing & $0.1$
    \end{tabularx}
    \caption{Optimizer hyper-parameters}
    \label{tab:optim}
\end{table}%

\subsection{Data}
\begin{figure*}
    \centering
    \includegraphics[width=0.6\textwidth]{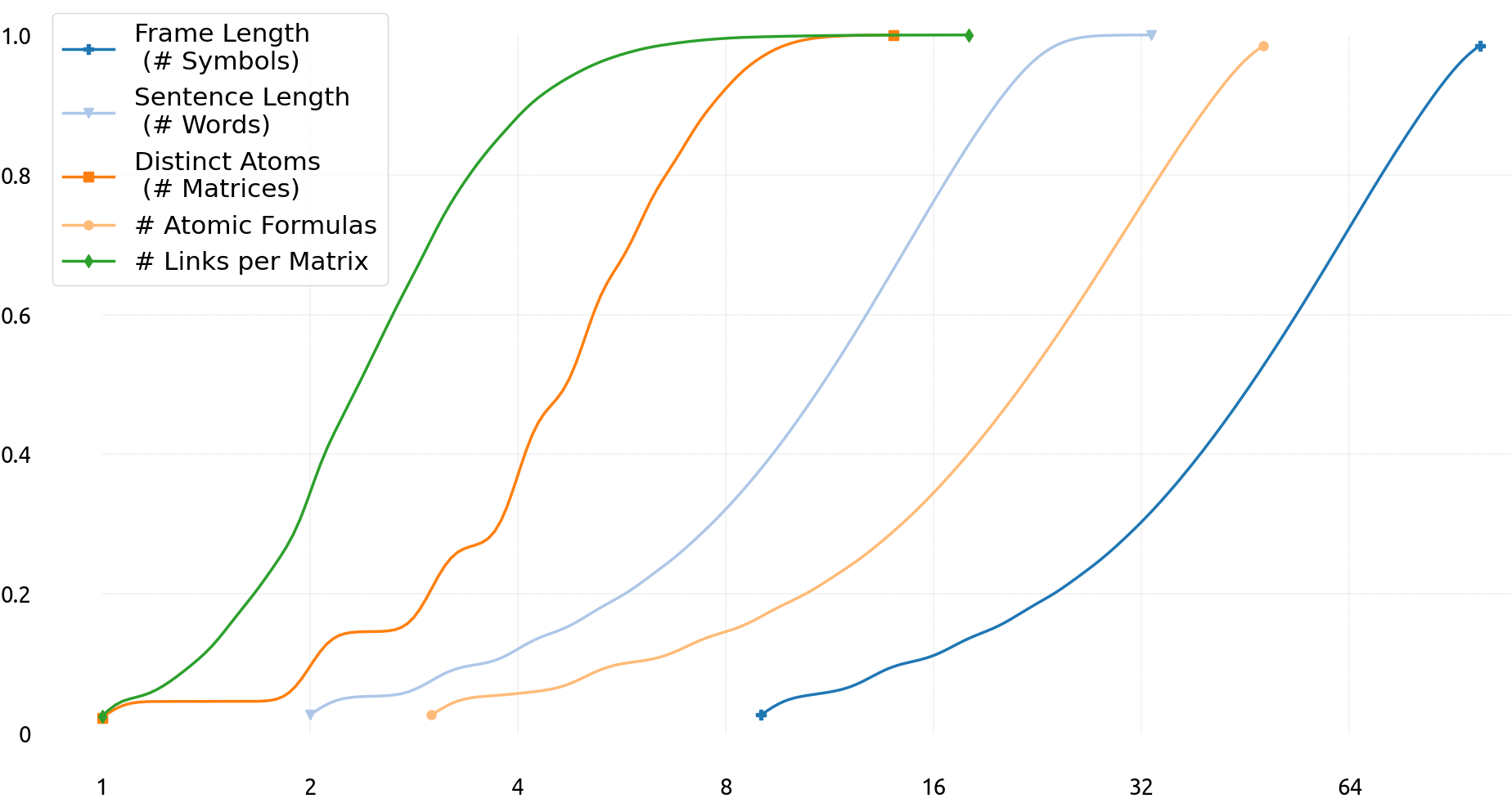}
    \caption{$\mathrm{log2}$-transformed cumulative distributions of symbol and word lengths, counts of atomic formulas, matrices and matrix sizes from the portion of the dataset trained on.}
    \label{fig:stats}
\end{figure*}

Figure~\ref{fig:stats} presents cumulative distributions of dataset statistics.
The kept portion of the dataset corresponds to roughly $97\%$ of the original, enumerating $55\,683$ training, $6\,971$ validation and $6\,957$ test samples.

\subsection{Performance}
Table~\ref{tab:lennum} summarizes the model's performance in terms of untyped term accuracy over the test set in the greedy setting, binned according to input sentence lengths.
Table~\ref{tab:Example} presents input-output pairs from sample sentences not included in the dataset.

\begin{table}[h]
    \centering
    \newcolumntype{L}{>{\raggedright\arraybackslash}p{0.375\columnwidth}}
    \begin{tabularx}{0.99\columnwidth}{@{}Lccc@{}}
    \textbf{Sentence Length} & \textbf{Total} & \textbf{Correct} & (\%)\\
    \midrule
    1 -- 5 & $808$ & $743$ & $92$ \\
    5 -- 10 & $1\,491$ & $1\,104$ & $74$ \\ 
    10 -- 15 & $1\,576$ & $919$ & $58$ \\
    15 -- 20 & $1\,206$ & $501$ & $42$ \\
    20 --  & $592$ & $154$ & $26$
    \end{tabularx}
    \caption{Test set model performance broken down by sentence length.}
    \label{tab:lennum}
\end{table}%

\begin{sidewaystable*}
    \centering
    \small{
    \newcolumntype{L}{>{\raggedright\arraybackslash}X}
    \newcolumntype{R}{>{\raggedleft\arraybackslash}X}
    \newcommand{\ra}[1]{\renewcommand{\arraystretch}{#1}}
    \ra{1.1}
    \begin{tabularx}{0.99\textwidth}{@{}L@{}}
    \toprule
    \textit{De voorafgaande stukjes over Wiskundige Omgangstaal hadden het vooral over het samenspel tussen woorden en formules.} \\
    ``The preceding articles on the Mathematical Vernacular mainly focused on the interplay between words and formules.'' \\
    \scriptsize{
    $
    ((\term{hadden}::\pp\lis\pron\lis\s \ 
    ((\term{vooral}::\pp\lis\pp)_\text{mod} \ 
    (\term{over}::\np\lis\pp \ 
    ((\term{tussen}::\ww\lis\np\lis\np \
    ((\term{en}::\np\lis\np\lis\np \ 
    (\term{woprden}::\np)^\text{cnj}) \ 
    (\term{formules}::\np)^\text{cnj})^\term{obj})_\text{mod} \ 
    ((\term{het}::\n\lis\np)_\text{det} \ 
    \term{samenspel}::\n))^\text{obj}))^\text{pc} \ 
    (\term{het}::\pron)^\text{obj}) \ 
    ((\term{over}::\np\lis\np\lis\np \ 
    (\term{Wiskundige\_Omgangstaal}::\np)^\text{obj})_\text{mod} \ 
    ((\term{voorafgaande}::\np\lis\np)_\text{mod} \ 
    ((\term{De}::\n\lis\np)_\text{det} \ 
    \term{stukjes}::\n)))^\text{su}
    $
    }\\
    \midrule
    \midrule
    \textit{In het wiskundig Nederlands worden vaak dezelfde fouten gemaakt als in het gewone Nederlands.} \\
    ``The same mistakes are often made in mathematical Dutch as in common Dutch.''\
    \scriptsize{
    $
    (\term{worden}::\ppart\lis\np\lis\s \ 
    ((\term{vaak}::\ppart\lis\ppart)_\text{mod} \ 
    ((\term{In}::\np\lis\ppart\lis\ppart \ 
    ((\term{wiskundig}::\np\lis\np)_\text{mod} \
    ((\term{het}::\n\lis\np)_\text{det} \ 
    \term{Nederlands}::\n))^\text{obj})_\text{mod} \ 
    \term{gemaakt}::\ppart))^\text{vc}) \ 
    ((\term{dezelfde}::\CP\lis\n\lis\np \ 
    (\term{als}::\pp\lis\CP \ 
    (\term{in}::\np\lis\pp \ 
    ((\term{gewone}::\np\lis\np)_\text{mod} \ 
    ((\term{het}::\n\lis\np)_\text{det} \ 
    \term{Nederlands}::\n))^\text{obj})^\text{cmp\_body})^\text{obcomp})_\text{det} \ 
    (\term{fouten}::\n)^\text{su}
    $
    } \\
    \midrule
    \midrule
    \textit{In het wiskundige taalgebruik is er meestal een scheiding aan te brengen tussen de echte wiskundige taal en de taal waarmee we over die wiskundige taal of over het wiskundige bedrijf spreken.} \\ 
    ``In mathematical discourse, there is usually a distinction to be made between the real mathematical language and the language with which we speak about the mathematical language or about the mathematical practice.`` \\ 
    -- \\
    \midrule
    \midrule
    \textit{Probeer zinnen steeds zo te stellen dat ze alleen op de door de schrijver bedoelde wijze zijn terug te lezen.} \\ 
        ``Try to always formulate sentences in such a way that they can only be read in the manner intended by the author.'' \\  -- \\ 
    \midrule
    \midrule
    \textit{In het Nederlands kunnen vele zinnen wat volgorde betreft omgegooid worden.}\\
    ``In Dutch, many sentences can be restructured as far as order is concerned.'' \\
    \scriptsize{
    $
    (\term{kunnen}::\Inf\lis\np\lis\s \ 
    (\term{worden}::\ppart\lis\Inf \ 
    ((\term{wat}::(\pron\lis\ssub)\lis\ppart\lis\ppart \ 
    \lambda x_0^\text{su}.
    ((\term{betreft}::\n\lis\pron\lis\ssub \ 
    (\term{volgorde}::\n)^\text{obj}) \ 
    x_0)^\text{rel\_body})_\text{mod} \ 
    ((\term{In}::\np\lis\ppart\lis\ppart \ 
    ((\term{het}::\n\lis\np)_\text{det} \ 
    \term{Nederlands}::\n)^\text{obj})_\text{mod} \ 
    \term{omgegooid}::\ppart))^\text{vc})^\text{vc}) \ 
    ((\term{vele}::\np\lis\np)_\text{mod} \ 
    \term{zinnen}::\np)^\text{su}
    $
    } \\
    \midrule
    \midrule
    \textit{In het Nederlands kunnen vaak twee zinnen tot \`{e}\`{e}n kortere worden samengetrokken.} \\ 
    ``In Dutch, two sentences can often be merged into a shorter one.'' \\ 
    \scriptsize{
    $
    (\term{kunnen}::\Inf\lis\np\lis\s \ 
    (\term{worden}::\ppart\lis\Inf \ 
    ((\term{vaak}::\ppart\lis\ppart)_\text{mod} \ 
    ((\term{In}::\np\lis\ppart\lis\ppart \ 
    ((\term{het}::\n\lis\np)_\text{det} \ 
    \term{Nederlands}::\n)^\text{obj})_\text{mod} \ 
    (\term{samengetrokken}::\pp\lis\ppart \ 
    (\term{tot}::\np\lis\pp \ 
    ((\term{\grave{e}\grave{e}n}::\adj\lis\np)_\text{det} \ 
    \term{kortere}::\adj)^\text{obj})^\text{ld}))^\text{vc})^\text{vc}) \ 
    ((\term{twee}::\n\lis\np)_\text{det} \ 
    \term{zinnen}::\n)^\text{su}
    $
    } \\ 
    \midrule
    \midrule
    \textit{Populaire taal is vaak minder beveiligd tegen dubbelzinnigheid dan nette taal, en het mengsel van beide talen is n\`{o}g gevaarlijker.} \\ 
    ``Informal language is often less protected against ambiguity than formal language, and the mixture of both languages is even more dangerous.'' \\ 
    \scriptsize{
    $
    (\term{en}::\s\lis\s\lis
    ((\term{is}::\ppart\lis\np\lis\s \ 
    ((\term{minder}::\CP\lis\ppart\lis\ppart \ 
    (\term{dan}::\np\lis\CP \ 
    ((\term{nette}::\np\lis\np)_\text{mod} \ 
    \term{taal}::\np)^\text{cmp\_body})^\text{obcomp})_\text{mod} \ 
    ((\term{vaak}::\ppart\lis\ppart)_\text{mod} \ 
    (\term{beveiligd}::\pp\lis\ppart \ (
    (\term{tegen}::\np\lis\pp \ 
    (\term{dubbelzinnighead}:\np)^\text{obj})^\text{pc})))^\text{vc} \ 
    ((\term{Populaire}::\np\lis\np)_\text{mod} \ 
    \term{taal}::\np)^\text{su})^\text{cnj}) \ 
    ((\term{is}::\type{ap}\lis\np\lis\s \ 
    ((\term{n\grave{o}g}::\type{ap}\lis\type{ap})_\text{mod} \ 
    \term{gevaarlijker}::\type{ap})^\text{predc} \ 
    ((\term{van}::\np\lis\np\lis\np \ 
    ((\term{beide}::\n\lis\np)_\text{det} \ 
    \term{talen}::\n)^\text{obj})_\text{mod} \ 
    (\term{het}::\n\lis\np)_\text{det} \ 
    \term{mengsel}::\n))^\text{su})^\text{cnj}
    $
    } \\ 
    \bottomrule
    \end{tabularx}
    }
    \caption{Greedy parses of the opening sentences of the first seven paragraphs of~\citet{de1979wiskundigen}, in the form of type- and dependency-annotated $\lambda$ expressions. Two of them (3 \& 4) yield no valid proof net; the remaining five are both valid and correct.}
    \label{tab:Example}
\end{sidewaystable*}

\end{document}